\documentclass[runningheads]{llncs}

\usepackage{graphicx}
\usepackage{amsmath}
\usepackage{color}
\usepackage{wrapfig}
\usepackage{adjustbox}
\usepackage{multirow}
\usepackage{makecell}

\usepackage{parskip}
\newcommand{\norm}[1]{\left\lVert#1\right\rVert}

\usepackage{hyperref}
\hypersetup{
    colorlinks=true,
    linkcolor=blue,
    filecolor=magenta,      
    urlcolor=magenta,
}

\begin{document}
%
\title{Federated Simulation for Medical Imaging}

%
%
\author{\ Daiqing Li$^{1} \thanks{Correspondence to \email{\{daiqingl,sfidler\}@nvidia.com}}$ \and
Amlan Kar$^{1,2,5}$ \and
Nishant Ravikumar$^{3}$ \and
Alejandro F Frangi$^{3,4}$ \and
Sanja Fidler$^{1,2,5}$}


\authorrunning{D. Li et al.}
%

\institute{$^1$ NVIDIA \hspace{1.5mm}
$^2$ University of Toronto \hspace{1.5mm}
$^3$ University of Leeds \hspace{1.5mm} \\
$^4$ KU Leuven \hspace{1.5mm}
$^5$ Vector Institute \hspace{1.5mm}
}

\maketitle              
\begin{abstract}
Labelling data is expensive and time consuming  especially for domains such as medical imaging that contain volumetric imaging data and require expert knowledge. Exploiting a larger pool of labeled data available across multiple  centers, such as in federated learning, has also seen limited success since current deep learning  approaches do not generalize well to images acquired with scanners from different manufacturers. We aim to address these problems in a common, learning-based image simulation framework which we refer to as \emph{Federated Simulation}. We introduce a physics-driven generative approach that consists of two learnable neural modules: 1) a module that synthesizes 3D cardiac shapes along with their materials, and 2) a CT simulator that renders these into realistic 3D CT Volumes, with annotations.
Since the model of geometry and material is disentangled from the imaging sensor, it can effectively be trained across multiple medical centers. We show that our data synthesis framework improves the downstream segmentation performance on several datasets. Project Page: \url{https://nv-tlabs.github.io/fed-sim/} .

\end{abstract}
\section{Introduction}
High quality pixel-level annotations, necessary for training fully supervised segmentation approaches, are prohibitively expensive to source for medical imaging data especially in the context of 3D volumes. This is due to the high dimensionality of the data and the complexity of the task of identifying tissue boundaries and manually delineating the object(s) of interest. Furthermore, identifying regions of interests often requires expert knowledge. 

An appealing alternative to labeling data, is to \emph{synthesize} it~\cite{kar2019meta,unberath2015,zhang20}. 
Generating synthetic data sets to learn from, by simulating medical images, has been proposed in several previous studies. This includes simulation of cardiac CTs \cite{unberath2015} from mesh-based parametric representations of cardiac shape. Among these, Unberath et al.~\cite{unberath2018deepdrr} were the first to use deep learning to estimate X-ray scatter and simulate digitally reconstructed radiographs (DRRs) from annotated CT volumes. 
~\cite{zhang20} employed unpaired image-to-image style transfer to further improve the similarity between real X-ray images and the DRRs. These methods focus on synthesizing DRRs or CTs conditioned on (given) annotations, but cannot produce novel shapes and annotations from the data distribution. 
Furthermore, these methods work at the pixel-level and need to be re-trained per modality. We, instead, break the synthesis process into both, synthesizing novel 3D shapes and materials, and physical sensor simulation allowing us to generate multiple imaging modalities along with their annotations.
We additionally aim to effectively exploit pools of data available across different acquistion sensors with as little as annotations as possible. We refer to this learning-based imaging simulation framework as `Federated Simulation'.

We introduce a physics-driven generative approach that consists of two modules: 1) a module that synthesizes 3D cardiac shapes along with their materials, and 2) a CT simulator that renders these into realistic CT volumes. Both are implemented as learnable neural network modules enabling us to simulate realistic cardiac CTs.
Since the model of geometry and material is disentangled from the imaging sensor, it can effectively be trained across different centers in a privacy preserving manner. Once trained, our model can synthesize a virtually infinite amount of data in a desired imaging modality. By design, our approach also produces ground-truth labels along with the CTs, enabling training of downstream machine learning models.
We showcase our data simulation framework to outperform the traditional federated learning approaches in our use case.

\section{Methodology}
\label{sec:method}

We aim to learn a generative model $S_\theta$, parametrized using neural networks, to synthesize CT volumes and their corresponding labels (in our case voxel segmentation labels). Here, $\theta$ are learnable parameters of our model that we learn from a given CT dataset $D$ with few annotated and several unlabeled volumes. We wish to learn $S_\theta$ such that it captures the \emph{essence} of the dataset $D$, and can generate new realistic samples from its distribution. These samples are then used as an auxilary labeled dataset for training downstream machine learning models (in our case a 3D segmentation neural network). 


We introduce our generative model in Sec.~\ref{ss:gen} and explain how we learn it for a single site in  Sec.~\ref{ss:learning}. Finally, in Sec.~\ref{ss:federated} we propose how to implement the algorithm in a federated setting across multiple data sites. 

\begin{figure}[t!]
  \centering
    \centering
    \includegraphics[width=0.98\linewidth]{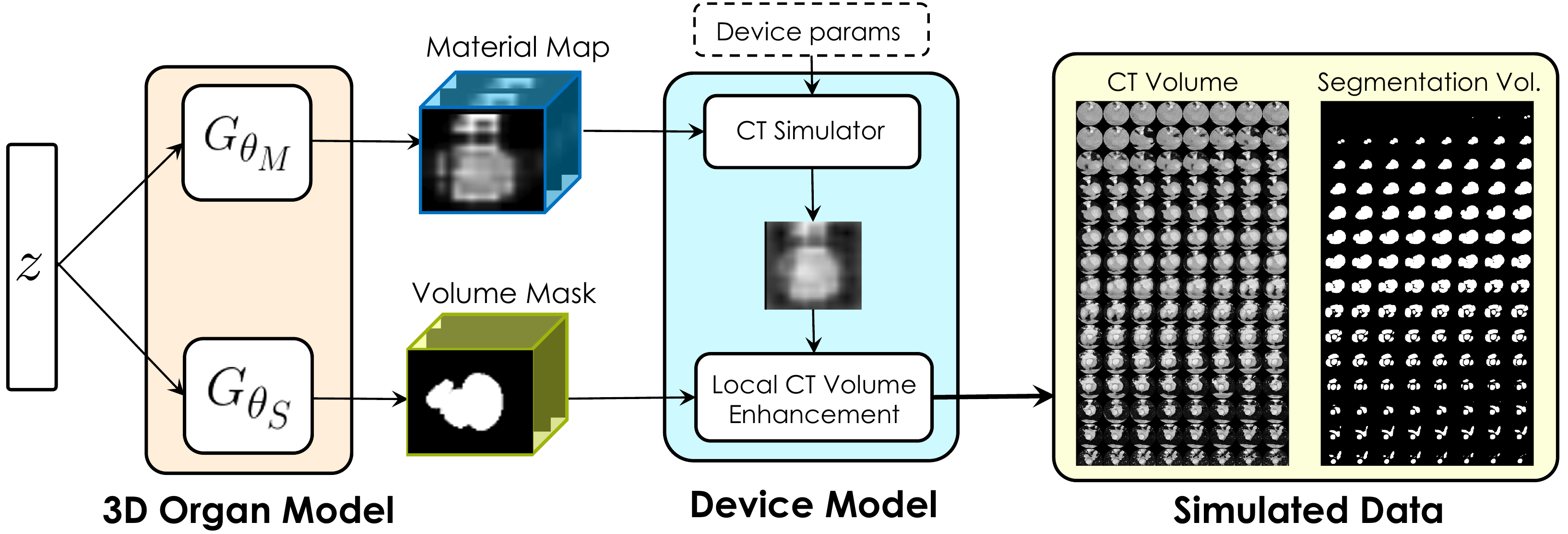}
    \caption{\footnotesize{\bf Our Generative Model:} We sample a latent vector from a normal distribution, and pass it through two neural networks to produce the organ's \emph{shape} and a \emph{material map}. These are then input to a differentiable CT renderer to produce a CT volume. A conditional GAN is then used to further improve the realism of the volume.}
    \label{fig:model}
\end{figure}

\subsection{Generative Model}
\label{ss:gen}
Our generative model (Fig.~\ref{fig:model})  generates an organ shape and a material map, both independent from an imaging device, that are then passed through a CT simulator to generate a synthetic CT volume with labels. 
Additionally, we \emph{enhance} the generated CT volume using a 
conditional Generative Adversarial Network (GAN)~\cite{park2019semantic} to further improve on realism. In the federated simulation setting, different sites jointly learn a global shape and material model, while each site maintains a site-specific GAN. Federated setting is discussed in details in Sec~\ref{ss:federated}.

\textbf{Shape/Material Generation:} From a latent vector $z \in \mathcal{Z}$, we aim to learn to generate a 3D organ shape and material properties around the shape. To ensure tractable learning, we constrain generated shapes to be physically plausible and provide control over the shape using a reduced set of parameters through a Statistical Shape Model (SSM). Thus, the shape parameter $\tau_S = G_{\theta_S}(z) \in \mathcal{R}^{21}$ operates on the SSM in order to generate a mesh of the organ. Along with a vector of SSM weights (in $\mathcal{R}^{14}$), we also generate a rigid transformation to be applied to the mesh (in $\mathcal{R}^7$, 3 rotation, 3 translation and 1 scale). In our case, the SSM is a parametric representation of the whole heart and its associated great vessel trunks (pulmonary artery and aorta), including seven regions, namely, the blood pool and myocardium of the left ventricle, the right ventricle, the left and right atria, and the vessels. We estimated our SSM using PCA~\cite{ravikumar17}(see supplementary material for details). We obtain the organ's mesh and convert it to a volumetric (voxel) representation for the CT simulator as $S= \text{voxelize}(B(\tau_S))$, with $B$ being our PCA basis.

From the same latent vector $z$, we also generate a coarse material voxel map $\tau_M = G_{\theta_M}(z) \in \mathcal{R}^{16\times16\times16}$. The material map is a combined representation of the voxel-wise tissue-properties, energy-dependent linear attenuation coefficient and material density at each point in the object. See supplementary material for implementation details of $G_{\theta_S}$ and $G_{\theta_M}$.

\textbf{CT Simulation:} The generated shape $S$ and material map $\tau_M$ are passed through a physics-based CT renderer to generate a voxelized label map $Y_z \in \mathcal{R}^{128\times128\times128}$ and a CT voxel volume $\tilde{X_z} \in \mathcal{R}^{16\times16\times16}$. Note that the generated CT volume is coarse due to the coarseness of the material map.

We use PYRO-NN~\cite{syben2019pyro} as our CT renderer, a python-based CT reconstruction framework which provides cone-beam forward and back-projection operations embedded as Tensorflow layers, enabling easy integration of the renderer within our generative neural network. As highlighted in~\cite{syben2019pyro} the forward and back-projection operators are differentiable, thus gradients can be efficiently propagated for end-to-end training of CT reconstruction networks. 


\textbf{CT Volume Enhancement:} CT-Images are dependent on factors such as the scanning machine and acquisition protocol at a site, which are not all modelled by our CT simulator. Additionally, our generated CT volume is a coarse representation. We thus use a GAN to both enhance and translate our simulated CT slices to look similar to target images, bridging the gap between simulation and real data. The generated coarse CT volume $\tilde{X}_z$ and label map $Y_z$ are used to generate the final high resolution synthetic CT slices. Specifically, we utilize GauGAN~\cite{park2019semantic} $(G_{\text{GAN}})$ to take a slice $k$ of $\tilde{X}_{z}$, $Y_z$ and slice index $k$ as input and produce a final CT slice image $X_{z,k}$, with label $Y_{z,k}$. We choose GauGAN since it is designed to respect semantic shape input, which is the label map in our case. Particularly, 1) we take the $k^{\text{th}}$ slice of $\tilde{X}_z$ and trilinearly upsample it to $\mathcal{R}^{128\times128}$, 2) take the $k^{\text{th}}$ slice of $Y_z$ and 3) create a $128\times128$ slice with a constant value of $\frac{k}{128}$ and concatenate them together as the input to $G_{\text{GAN}}$.

\textbf{Complete generative process} can be written succinctly as,
\begin{align*}
    S_z &= \text{SSM}(G_{\theta_S}(z))\\
    \tilde{X}_z, Y_z &= \text{CT}_{\text{sim}}(S_z, G_{\theta_M}(z))\\
    X_{z,k} &= G_{\text{GAN}}(\tilde{X}_{z,k}, Y_{z,k}, k) \quad \forall \: k
\end{align*}
where $S_z$ represents the shape obtained from the SSM with parameters $G_{\theta_S}(z)$.

\subsection{Learning}
\label{ss:learning}
We train our generative model using the Generative Latent Optimization (GLO)~\cite{bojanowski2017optimizing} framework in two stages: pre-training and unsupervised training. First, we pre-train the model using the labeled training subset, and then fine-tune the model in a semi-supervised fashion using the rest of the unlabelled training data. We first introduce the GLO framework and then describe our training stages.

\textbf{GLO~\cite{bojanowski2017optimizing}:} GLO is a technique for learning generative networks using only reconstruction losses. In our case, every volume in the training set is coupled with a particular latent vector $z^i$ (initialized from a unit normal distribution), which is simultaneously learnt along with a generation process that transforms $z^i$ into a volume. Learning is done by optimizing reconstruction losses between the generated volume and the corresponding ground truth volume. The set of learnt latent vectors $\{z^i\}$ can be fit to a parametric distribution (\emph{eg.} multivariate normal), which is sampled from to generate new volumes. We choose GLO due to its stability in training while enjoying visual-appealing samples property of GANs. We also found that GLO has better sample quality in small-data regime comparing to Meta-Sim\cite{kar2019meta}.

\textbf{Pre-training:} We use the training CT volumes which have ground truth annotations to pretrain the parameter generation module ($G_{\theta_S}$ and $G_{\theta_M}$). We use the mean Intersection-Over-Union (mIoU) metric to learn $G_{\theta_S}$ (in our case, generated $Y_z^i$ and ground-truth $Y^i$ are binary), and a combination of mean-square-error and perceptual loss~\cite{johnson2016perceptual} using all VGG-19~\cite{simonyan2014very} layer features for $G_{\theta_M}$.
\begin{align*}
    L_{\text{IoU}}(\theta_S) &= 1 - \frac{Y_z^i \odot Y^i}{Y_z^i + Y^i - Y_z^i \odot Y^i}\\
    L(\theta_M) &= \norm{\tilde{X}_z^i - \tilde{X}^i}_2^2 + L_{\text{perc}}(\tilde{X}_z^i, \tilde{X}^i)
\end{align*}
where $\tilde{X}^i \in \mathcal{R}^{16\times16\times16}$ is the trilinearly downsampled real CT volume $X^i$ , $\odot$ represents the hadamard product and other operations are element-wise for matrices. $L_\text{perc}$ is implemented following~\cite{park2019semantic}.  Backpropagation through the SSM is done by calculating the gradients using finite-differences on the low-dimensional SSM parameters.

\begin{figure}[t!]
  \centering
  \begin{minipage}{0.55\textwidth}
    \centering
    \includegraphics[width=0.98\linewidth]{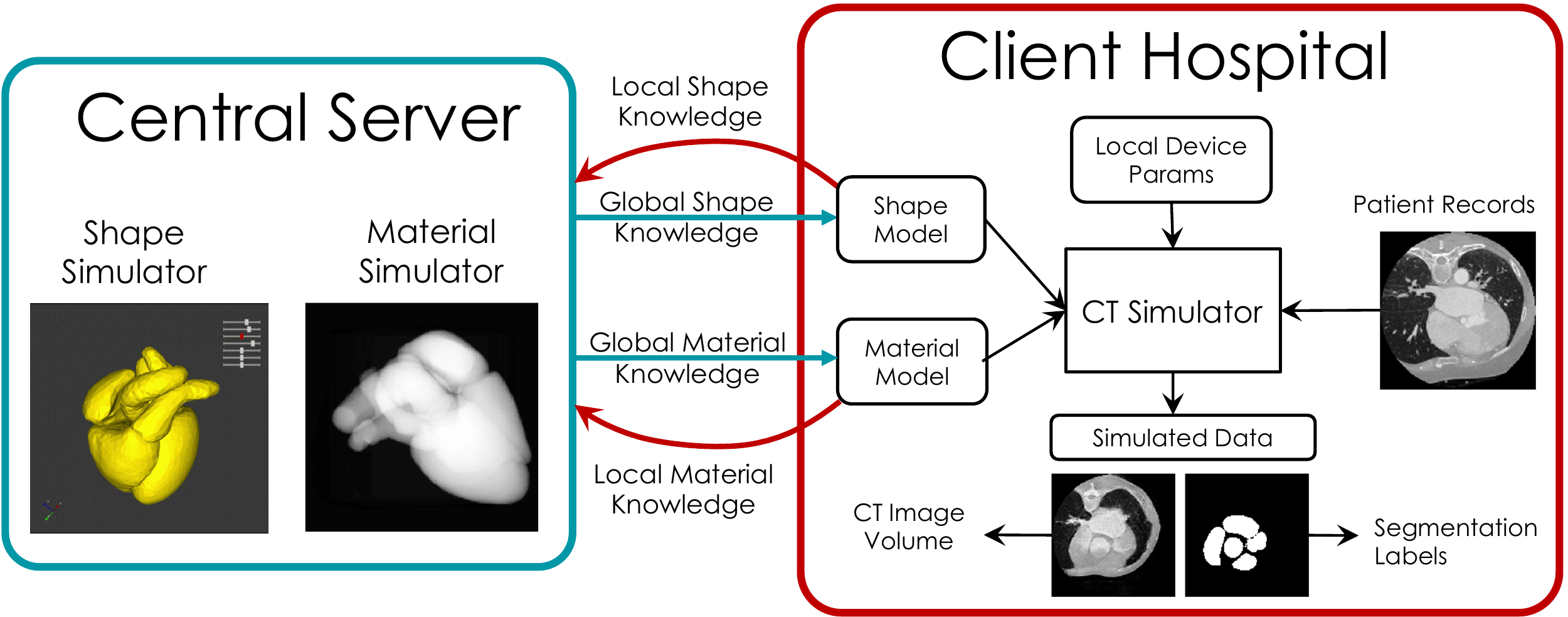}
  \end{minipage}
  \begin{minipage}{0.44\textwidth}
    \caption{\footnotesize{\bf Federated Simulation:} Central server trains a generative model of 3D organ shape and material, which is deployed to centers. Clients train local device models that mimic their sensors. Clients send gradients w.r.t. shape and material back to server.}
    \label{fig:federated}
  \end{minipage}    
\end{figure}

After pre-training $G_{\theta_S}$ and $G_{\theta_M}$, we pre-train our conditional GAN $G_{\text{GAN}}$. Specifically, in this phase, we use a slice $k$ of the ground truth labels $Y^i$ instead of the generated labels $Y_z^i$, and the generated material $\tilde{X}_z^i$ as input to the GAN, and supervise it with the associated real CT Volume $X^i$, using the same loss functions as~\cite{park2019semantic}. Using the ground truth label ensures that the input label and the output image have the same exact shape, resulting in the GAN learning to generate a high-resolution CT that respects the input shape.

\textbf{Semi-Supervised Learning:} We utilize the  unlabelled data (which is typically more widely available) in the training set to improve our simulation. The abundance of unlabelled data and the cost of annotation makes this a compelling proposition. In this stage, we alternate between training on supervised data (explained above), and training on unsupervised data. We first fit a multivariate normal distribution to the latent vectors optimized in the pre-training phase, and sample new random latent vectors $z^i$ for the new unlabelled data-points from this distribution. To train unsupervised, we run our full generative process from every $z^i$ to generate a high resolution CT image $X_{z,k}^i$ for some random slice index $k$ from $z^i$ and use the same loss function for the GAN as above to train. This phase adapts the model to be able to learn shape and material properties for data points without ground-truth segmentation annotations. Note that we freeze the GAN discriminator in this phase, and observe that it strongly improves training stability.

\subsection{Federated Simulation}
\label{ss:federated}
Medical data (CTs in our case) is usually available at multiple sites, each with their specific acquisiton parameters and privacy concerns, which makes both, consolidating data and training on consolidated data (domain adaptation) difficult. With our learning-based simulation, we demonstrate \emph{federated simulation}, where we learn our generative model with data from multiple hospitals in a federated fashion. Because of our disentanglement of shape and material from the CT process, we are in an advantageous position where we can learn shape and material parameters across multiple sites, and \emph{render} generated shapes and materials from this global distribution into CT volumes from a particular hospital's distribution through the CT simulator and the site-specific (local) enhancement GAN. This mitigates both issues of data consolidation and domain adaptation.

Fig.~\ref{fig:federated} depicts this process. We learn one $G_{\theta_S}$ and $G_{\theta_M}$ to model the distribution of shapes and materials across sites, and learn a site-specific $G_\text{GAN}$. In every case, we take one step update from each site, accumulate gradients and run a step of gradient descent at the server, and broadcast the updated weights $\theta_S$ and $\theta_M$ back to the sites. Note that this is a simple federated setting and serves to demonstrate our method; but a real deployment would require additional engineering considerations. See supplementary material for details.
\section{Experiments}
\label{sec:exp}
We validate our method on three Cardiac-CT datasets on both the single hospital train/test and our federated scenario. Additional ablation studies are conducted to validate our design choices. We split each dataset into train, validation and test subsets and experiment with different sizes of labels made available in the training set. All experiments are done using \emph{five-fold cross validation}, and we report the mean and std. deviation of the test set performance.

\begin{table}[h]

\begin{minipage}{0.485\linewidth}
\caption{Dataset Split Sizes}
\centering
\begin{small}
\begin{tabular}{|c|c|c|c|}
    \hline 
     Dataset & Train & Val & Test\\ \hline\hline
     CT20~\cite{zhuang16} & 12 & 4 & 4\\ \hline
     CT34LC~\cite{xu2019whole} & 20 & 7 & 7\\ \hline
     CT34MC\cite{xu2019whole} & 20 & 7 & 7\\ \hline
\end{tabular}
     \end{small}

\label{tab:datasets}
\end{minipage}
\begin{minipage}{0.485\linewidth}
\caption{Ablation Studies}
\centering
\begin{tabular}{|l|l|}
\hline
Method & Perf. \\ \hline \hline
Rand Shape + GAN          & 67.07 \\ \hline
+ Alpha blend         & 68.80 \\ \hline
+ Poisson blend       & 76.44 \\ \hline
Ours Fix-Mat & \textbf{78.92} \\ \hline
\end{tabular}

\label{tab:ablate}
\end{minipage}
\end{table}

\textbf{Datasets}: We will refer to our datasets as \textbf{CT20}, \textbf{CT34LC} and \textbf{CT34MC}. CT20 data was collected from healthy adults at the Shanghai Shugang Hospital, China, and provided as part of the MM-WHS challenge \cite{zhuang16}. The CT34LC and CT34MC data~\cite{xu2019whole} were collected from congenital heart disease patients, whose ages ranged from 1 month-21 years. We split it into two equal-sized datasets based on pathological differences. Tab.~\ref{tab:datasets} details the split sizes for each of the three datasets. Large difference in age demographics and the presence/absence of pathology correspond to substantial variations in cardiac shape pose a significant challenge for generalization in learning.  

\textbf{Evaluation Metric:} To evaluate the performance of our generated dataset, we train a 3D-Unet~\cite{cciccek20163d} for binary segmentation of the heart region and measure its performance on the respective test set. Such task-based performance evaluation of generative models has been proposed in~\cite{bass2019image}, and we adopt it here. For every experiment with synthetically generated data, we evaluate by first pre-training on the synthetic data and then fine-tuning on the available real data. We note that semi-supervised training techniques~\cite{lee2013pseudo} could be used for the segmentation model to make use of unlabelled CT-volumes in all cases, which we omit here. 

\begin{figure}[t!]
\centering
\begin{minipage}{0.485\linewidth}
\begin{tabular}{ccccc}
\rotatebox{90}{\ \ \,{\color{black}{\scriptsize CT20}}} &\includegraphics[width=0.225\linewidth]{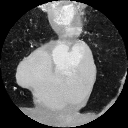}
&
\includegraphics[width=0.225\linewidth]{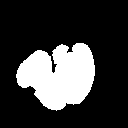}
&
\includegraphics[width=0.225\linewidth]{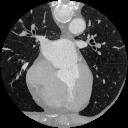}
&
\includegraphics[width=0.225\linewidth]{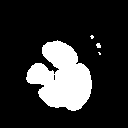}\\

\rotatebox{90}{\ {\color{black}{\scriptsize CT34LC}}} &\includegraphics[width=0.225\linewidth]{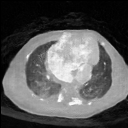}
&
\includegraphics[width=0.225\linewidth]{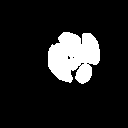}
&
\includegraphics[width=0.225\linewidth]{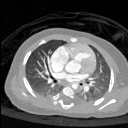}
&
\includegraphics[width=0.225\linewidth]{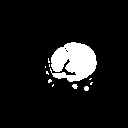}\\

\rotatebox{90}{\,{\color{black}{\scriptsize CT34MC}}} &\includegraphics[width=0.225\linewidth]{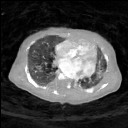}
&
\includegraphics[width=0.225\linewidth]{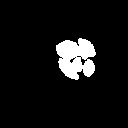}
&
\includegraphics[width=0.225\linewidth]{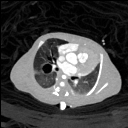}
&
\includegraphics[width=0.225\linewidth]{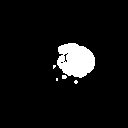}\\
\end{tabular}
\end{minipage}
\hspace{1mm}
\begin{minipage}{0.485\linewidth}
\begin{tabular}{ccccc}
\rotatebox{90}{\ \ \,{\color{black}{\scriptsize CT20}}} &\includegraphics[width=0.225\linewidth]{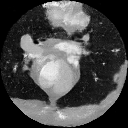}
&
\includegraphics[width=0.225\linewidth]{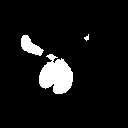}
&
\includegraphics[width=0.225\linewidth]{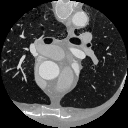}
&
\includegraphics[width=0.225\linewidth]{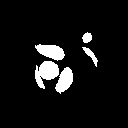}\\

\rotatebox{90}{\ {\color{black}{\scriptsize CT34LC}}} &\includegraphics[width=0.225\linewidth]{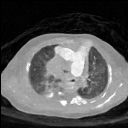}
&
\includegraphics[width=0.225\linewidth]{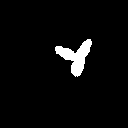}
&
\includegraphics[width=0.225\linewidth]{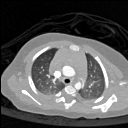}
&
\includegraphics[width=0.225\linewidth]{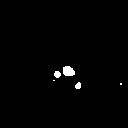}\\

\rotatebox{90}{\,{\color{black}{\scriptsize CT34MC}}} &\includegraphics[width=0.225\linewidth]{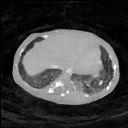}
&
\includegraphics[width=0.225\linewidth]{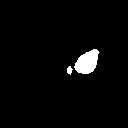}
&
\includegraphics[width=0.225\linewidth]{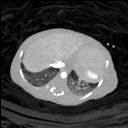}
&
\includegraphics[width=0.225\linewidth]{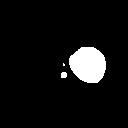}\\
\end{tabular}
\end{minipage}
\caption{Qualitative Results: Two sets of examples per dataset. First two columns show random samples (one slice) from our full model on each of the datasets. Last two columns show nearest neighbour (same slice) from the training set. Our model can generate plausible yet novel data samples with annotations.}
\label{fig:qual}
\end{figure}

\textbf{Single Site Simulation:} We first evaluate our model independently per dataset. We compare against training a supervised model using the subset of training dataset with labels (\textbf{Lower Bound}) and training the supervised model on the full training dataset (\textbf{Upper Bound}) using extra training labels that our method does not have access to. We also compare three variants of our model, 1) instead of predicting a material map, using a fixed atlas of attenuation coefficients~\cite{unberath2018deepdrr} (\textbf{Ours Fix-Mat}) (in this setting, the material map and the rendering is already high-resolution), 2) using only our pre-trained method (\textbf{Ours Pre}) and 3) using our full method with semi-supervised training (\textbf{Ours-Full}). We also evaluate with different amounts of labels available from the training set. 20 synthetic volumes are generated when using our generative model to synthesize data, and all 3D-Unet training was done with a batch size of one. Tab.~\ref{tab:hosp} summarizes these results, and shows the effect of learning material parameters, as well as utilizing unlabelled data to learn the simulator. We see that our method beats the lower bound across all datasets, sometimes even beating the upper bound, meaning that using our synthesized data improves performance more than using all annotations from the training set. Some generated samples per dataset are shown in Fig.~\ref{fig:qual}. In the figure, we also show that our method generates novel samples by showing the nearest neighbour sample (computed on the whole CT-Volume) from the training set. 

\begin{table}[t!]
\caption{Quantitative Results of training a Unet-3D binary segmentation model on our generated data on three datasets. We see that data generated by our methods (with access to a small subset of training labels) outperforms baselines on both the single site and federated simulation case. It sometimes outperforms the upper bound of using the full training set as well. Method with highest mean is in bold.}
 \begin{adjustbox}{width=\textwidth}
\begin{tabular}{p{5.8mm}|l|l|l|l|l|l|l|}

\cline{3-8} \multicolumn{2}{c|}{}&
  \multicolumn{2}{c|}{CT20} &
  \multicolumn{2}{c|}{CT34LC} &
  \multicolumn{2}{c|}{CT34MC} \\
\cline{3-8} \multicolumn{2}{c|}{} & Label Sz. 4 & Label Sz. 8 & Label Sz. 6 & Label Sz. 12 & Label Sz. 6 & Label Sz. 12 \\ \hline \cline{2-8}
\multirow{5}{*}{\rotatebox{90}{\ {\color{black}{\makecell{\scriptsize{Single Site$\ \ \  $}\\[-1.2mm]\scriptsize{Simulation$\ \ \,\, $}}}}}} & Lower Bound                                                      & 87.65$\pm$2.20 & 86.33$\pm$2.09 & 85.25$\pm$4.57 & 87.27$\pm$2.52  & 85.32$\pm$1.50 & 84.65$\pm$2.23  \\ \cline{2-8}
& Ours-Fix-Mat &87.87$\pm$4.28 &88.32$\pm$5.11 &86.33$\pm$1.69 &88.32$\pm$1.81 &83.13$\pm$2.19 &84.91$\pm$1.97 \\ \cline{2-8}
& Ours-Pre &87.78$\pm$4.28 &91.35$\pm$1.55 &\textbf{87.39$\pm$2.34} &88.01$\pm$1.30 &\textbf{85.92$\pm$1.37} &84.91$\pm$1.35 \\ \cline{2-8}
& Ours-Full &\textbf{88.95$\pm$2.97} &\textbf{91.39$\pm$1.27} &85.91$\pm$3.61 &\textbf{88.98$\pm$1.86} &84.79$\pm$1.73 & \textbf{85.13$\pm$2.31} \\ \cline{2-8}\noalign{\vskip\doublerulesep
         \vskip-\arrayrulewidth}\cline{2-8}
& Upper Bound & \multicolumn{2}{c|}{89.81 $\pm$ 2.50} & \multicolumn{2}{c|}{88.76 $\pm$ 1.82} & \multicolumn{2}{c|}{85.75 $\pm$ 2.43}   \\ \cline{2-8}\noalign{\vskip\doublerulesep
         \vskip-\arrayrulewidth}\hline

\multirow{3}{*}{\rotatebox{90}{\ \ \,{\color{black}{\makecell{\scriptsize{Fede.$\,$}\\[-1.2mm] \scriptsize{Sim}}}}}} 
& Direct-FL & 91.65$\pm$2.69 & 92.45$\pm$1.52 & 89.34$\pm$1.27 &   90.17$\pm$1.40 & 87.11$\pm$1.55 & 87.29$\pm$1.84  \\ \cline{2-8}
& Ours-Sim-FL                                                       &  90.71$\pm$1.62         &  \textbf{93.45$\pm$1.21}           &    88.33$\pm$2.10          &     \textbf{90.19$\pm$2.21}          &    \textbf{87.15$\pm$1.66}         &   \textbf{87.40$\pm$1.71}            \\ \hline
\end{tabular}
\end{adjustbox}

\label{tab:hosp}
\end{table}

\textbf{Federated Simulation:} Next, we evaluate our method in the Federated setting. Specifically, we  simulate the federated setting by using our three datasets as three different sites. We compare against training the segmentation network directly in a federated setting~\cite{mcmahan2016communication} (\textbf{Direct-FL}). In the federated simulation scenario (\textbf{Ours-Sim-FL}), each hospital generates data by sampling shapes and materials from the shared model and using their local GAN (see Fig.~\ref{fig:federated}). We argue that this can reduce domain-adaptation issues in sharing data (different devices, protocols etc.), since shapes and materials from other hospitals' distributions can be \emph{rendered} under another hospital's conditions using the CT Simulator and the site-specific GAN. The federated learning baseline (Direct-FL) averages three gradients from the sites, and thus runs at an effective batch size of three. Therefore, for fair reporting, all other 3D-Unet training also uses a batch size of three. Tab.~\ref{tab:hosp} presents these results. For all methods, we used the combination of lower label sizes (4,6,6) of all datasets for one experiment, and all the higher label sizes (8,12,12) for another. Every model was fine-tuned on the particular site's labelled training data before reporting. We see that using federated-simulation performs comparably or slightly worse with the federated baseline across all datasets, showing that we can indeed learn to simulate and share a simulator across sites instead of sharing/working on real data samples, which comes with significant privacy concerns.

\textbf{Performance on out-of-distribution samples:} We simulate the case where a patient from one hospital A goes to hospital B, by running the GAN trained for hospital B on the patient's GT segmentation mask and downsampled CT image. An ideal segmentation network would perform well on this out-of-distribution sample. See supplementary material for results.

\textbf{Ablation Studies:} We ablate our choice of learning shape parameters for the SSM in Tab.~\ref{tab:ablate}. These experiments use are performed in the \textbf{Ours-Fix-Mat} setting. Specifically, we compare with randomly generating shapes (instead of learning) from the SSM, alpha blending the heart (foreground) from the $\text{CT}_\text{sim}$ output with the background from the GAN output, poisson blending~\cite{perez2003poisson} in the same case and our method of learning the shape and using the full output of the GAN. These results show the efficacy of learning the shape parameters, and using the conditional GAN to generate the final enhanced output. All experiments here train \emph{only} on synthetic data and are trained on the CT20 dataset (with 4 training labels). 

\section{Conclusion}
In this paper, we introduced a generative model for synthesizing labeled cardiac CT volumes that mimic real world data. Our model abstracts modeling of the shape and material away from the imaging sensor, which enables it to learn in a federated setting, within a framework we call federated simulation. We show that using data generated by our method in both single-site and federated settings improves performance of a downstream 3D segmentation network. Our method currently is using a SSM to parameterize the shape which has limited representation ability. In the future work, we aim to explore a more flexible shape representation and extend the current framework to generate and learn from multiple image sensors (MR, CT etc.).

%
%
%
\bibliographystyle{splncs04}
\bibliography{mybibliography}

\end{document}


%
\title{Federated Simulation for Medical Imaging \\ Appendix}

%
%
%

\author{\ Daiqing Li$^{1} \thanks{Correspondence to \email{\{daiqingl,sfidler\}@nvidia.com}}$ \and
Amlan Kar$^{1,2,5}$ \and
Nishant Ravikumar$^{3}$ \and
Alejandro F Frangi$^{3,4}$ \and
Sanja Fidler$^{1,2,5}$}

\institute{$^1$ NVIDIA \hspace{1.5mm}
$^2$ University of Toronto \hspace{1.5mm}
$^3$ University of Leeds \hspace{1.5mm} \\
$^4$ KU Leuven \hspace{1.5mm}
$^5$ Vector Institute \hspace{1.5mm}
}


\authorrunning{D. Li et al.}
%

%
\maketitle              
\section{Out-of-Distribution Simulation Experiments}
 We simulate the case where a patient from one hospital A goes to hospital B, by running the GAN trained for hospital B on the patient's ground truth segmentation mask and downsampled CT image. An ideal segmentation network would perform well on this out-of-distribution sample. We summarize results in Tab.~\ref{tab:hosp}, where we see that our method consistently outperforms or performs similarly to our baseline methods on dealing with these (simulated) out-of-distribution inputs, which comes from our disentanglement of the sensor and content (shape and material), helping segmentation models trained on our simulated data generalize better. This experiment is in simulation since gaining such data (a patient with data at two different sites) is a challenge, but we hope to be able to perform this experiment on real data in the future.
 
\begin{table}[h!]
\caption{Quantitative Results of the Out-Of-Distribution simulation experiment, where we test performance on simulated data of patient from one hospital going to another hospital. Method with highest mean is in bold.}
 
\begin{adjustbox}{width=\textwidth}
\begin{tabular}{p{5.8mm}|l|l|l|l|l|l|l|}

\cline{3-8} \multicolumn{2}{c|}{}&
  \multicolumn{2}{c|}{CT20 Label Sz. 8} &
  \multicolumn{2}{c|}{CT34LC Label Sz. 12} &
  \multicolumn{2}{c|}{CT34MC Label Sz. 12} \\
\cline{3-8} \multicolumn{2}{c|}{} & CT34LC & CT34MC & CT20 & CT34MC & CT20 & CT34LC \\ \hline \cline{2-8}
\multirow{5}{*}{\rotatebox{90}{\ {\color{black}{\makecell{\scriptsize{Single Site$\ \ \  $}\\[-1.2mm]\scriptsize{Simulation$\ \ \,\, $}}}}}} & Lower Bound                                                      & 72.56$\pm$2.02 & 67.52$\pm$1.94 & 84.88$\pm$3.31 & 83.03$\pm$3.47  & 84.41$\pm$3.13 & 86.27$\pm$2.79  \\ \cline{2-8}
& Ours-Fix-Mat &73.96$\pm$3.12 &69.79$\pm$1.49
&\textbf{87.30$\pm$2.65} &\textbf{84.41$\pm$2.41} &84.14$\pm$2.62 &86.94$\pm$2.75  \\ \cline{2-8}
& Ours-Pre &\textbf{76.58$\pm$3.01} &\textbf{71.84$\pm$2.21} &86.22$\pm$2.20 &83.65$\pm$2.38 &81.54$\pm$2.01 &84.46$\pm$3.20 \\ \cline{2-8}
& Ours-Full &75.61$\pm$1.28 &70.55$\pm$0.77 &84.87$\pm$2.92 &84.08$\pm$3.11 &83.13$\pm$2.63 & 85.93$\pm$2.17 \\ \cline{2-8}\noalign{\vskip\doublerulesep
         \vskip-\arrayrulewidth}\cline{2-8}
& Upper Bound & 76.19$\pm$1.64& 71.69$\pm$1.91 & 84.01$\pm$2.54 &82.73$\pm$3.23 & 84.15$\pm$1.86
&86.39$\pm$2.63\\ \cline{2-8}\noalign{\vskip\doublerulesep
         \vskip-\arrayrulewidth}\hline

\multirow{3}{*}{\rotatebox{90}{\ \ \,{\color{black}{\makecell{\scriptsize{Fede.$\,$}\\[-1.2mm] \scriptsize{Sim}}}}}} 
& Direct-FL & \textbf{78.33$\pm$3.37} & 73.05$\pm$3.53 & 85.63$\pm$2.13 &   84.67$\pm$2.90 & 83.86$\pm$2.06 & \textbf{87.95$\pm$2.16}  \\ \cline{2-8}
& Ours-Sim-FL & 78.07$\pm$3.53 & \textbf{74.11$\pm$2.77}           &    \textbf{86.79$\pm$1.86}          &     \textbf{84.70$\pm$2.41}          &    \textbf{84.52$\pm$2.94}         &   87.76$\pm$2.83            \\ \hline
\end{tabular}
\end{adjustbox}

\label{tab:hosp}
\end{table}
\section{Shape Preprocessing Details}
Our shape model are obtained from MM-WHS challenge \cite{zhuang16} MRI annotation. We convert the 20 cardiac volume label into mesh using Marching Cube \cite{lorensen1987marching}. The extracted mesh model contains 4319 vertices and 8610 faces. We then estimate our SSM model using PCA to obtain the mean shape and the vectors of SSM weights. The shape class mean and covariance can be written as:
\begin{equation}
    \tilde{s} = \frac{1}{M}\sum_{i=1}^{M}s_i
\end{equation}
\begin{equation}
    C = \frac{1}{M-1}\sum_{i=1}^{M}(s_i-\tilde{s})(s_i-\tilde{s})^T
\end{equation}
The PCA of the shape produces $l$ eigenvectors $\Phi=[\varphi_{1}\varphi_{2}...\varphi_{l}]$ and the corresponding eigenvalues $\mathbf{\Lambda}=diag(\lambda_{1},\lambda_{2},...,\lambda_{l}])$. Then the new shape can be approximated from the following linear generative model:
\begin{equation}
    s \approx \tilde{s} + \Phi b
\end{equation}
where $b \in \mathcal{R}^{14}$ are shape parameters. In our experiment, we use the first 14 eigenvectors and we limite the range of $b$ from $-1.5\sqrt{\lambda}$ to $1.5\sqrt{\lambda}$.

\section{Shape/Material Parameter Generative Model Implementation Details}
In all our experiments, we choose our latent vector $z \in \mathcal{R}^{32}$. Our Generative Model of Shape $G_{\theta_S}$ is parametrized as a three layers Multilayer Perceptron(MLP). Each layer is a linear layer followed by an Leaky-ReLU activation except for the last layer where the activation function is Tanh. The layer weights are of size ${32\times256}$, ${256\times128}$, and ${128\times21}$. The design of our generative model of material $G_{\theta_M}$ is based on \cite{wu2016learning}. It consists of three fully convolutional layers with $\{256,128,1\}$ number of channels , kernel sizes of $\{3,3,3\}$ with a stride of $1$. Similar to \cite{wu2016learning}, we add batch normalization and ReLU layers between convolutional layers, and a Tanh layer at the end. Instead of using Transposed Convolutions to increase the output spatial dimension, we use Nearest-Neighbour Upsampling before each convolution to avoid checkerboard artifacts~\cite{odena2016deconvolution}. Specifically, we upsample with a scaling factor 4 to transform input $z$ with spatial dimension ${1\times1\times1}$ (and 32 channels) into ${4\times4\times4}$. Then we use two more upsampling layers with scaling factor 2 before each convolution layer to get the final output in $\mathcal{R}^{16\times16\times16}$.

\section{Semi-Supervised Learning Training Details}
At the beginning of this phase of training, we fit a multivariate normal  distribution  to  the  latent  vectors  optimized  in  the  pre-training  phase and  sample  new  random latent vectors $z_i$ for the new unlabelled data-points from this distribution. The intuition is that unlabelled data and labelled data come from the same data distribution, thus the latent representation of unlabelled data-point should come from the same latent distribution. While training, we use two Adam~\cite{kingma2014adam} optimizers with two different learning rate schedules for labelled and unlabelled data respectively. When training with labelled data (data used in the pre-training phase), we use a learning rate $1e^{-4}$ for $z_i$, $G_{\theta_S}$ and $G_{\theta_M}$ and $1e^{-5}$ for $G_{GAN}$. For the unlabelled data, we use a learning rate $1e^{-3}$ for $z_i$, $G_{\theta_S}$ and $G_{\theta_M}$ and $1e^{-4}$ for $G_{GAN}$. As the unlabelled data was not used in the pre-trained stage, we use a larger learning rate these data points. These learning rates are used for the first 30 epochs, and we linearly decay the learning rate of all the models to 0 for the last 30 epochs. Note that we freeze the weights of Discriminator during the training with the assumption that it can already distinguish fake/real data well by learning from labelled data, which we found this stabilize the training process significantly.

\section{Federated Learning Setup Details}
In our Federated Learning experiment, we setup the gradient communication between the clients and the server synchronously with a gradient update step 1. Specifically, we assume each client holds a private dataset (in our case, CT20, CT34\_LC and CT34\_MC) and the same model as in the server's site. At each training step in the client site, the clients will send back the gradient with respect to the current model sequentially to the server. In the server site, the server will do gradient aggregation and update the model parameters once it receives all the gradient from the clients. After the gradient update, the server will send back the new model to each of the client. This process continues until a maximum number of iteration. Implementation wise, this is equivalent to maintain a mini-batch of each client's private data and at each training step, the model will do a forward pass sequentially for each of the mini-batch. And then update the current model's parameters using the averaged accumulated gradient from all the mini-batches. In our experiment, we choose batch size one for each of the client, which makes the overall effective batch size three since we do one gradient update step after three forward passes. To compromise it, we also use batch size three for our method in FL setup. Note that in the FL baseline implementation, we use validation set from all the clients to select the best model and adjust the learning rate, which is different from our methods where we use fix number of iteration to pick the model. This difference might give our FL baseline advantage since it can learn a more generalized model using average score from all validation set. Under this setup, our model shows comparable performance comparing to the baseline. We will further exploit different options and a more realistic FL implementation for future work.

\begin{figure}[t!]
\centering
\begin{tabular}{ccccc}
\rotatebox{90}{\ \ \,{\color{black}{\scriptsize CT20}}} &\includegraphics[width=0.225\linewidth]{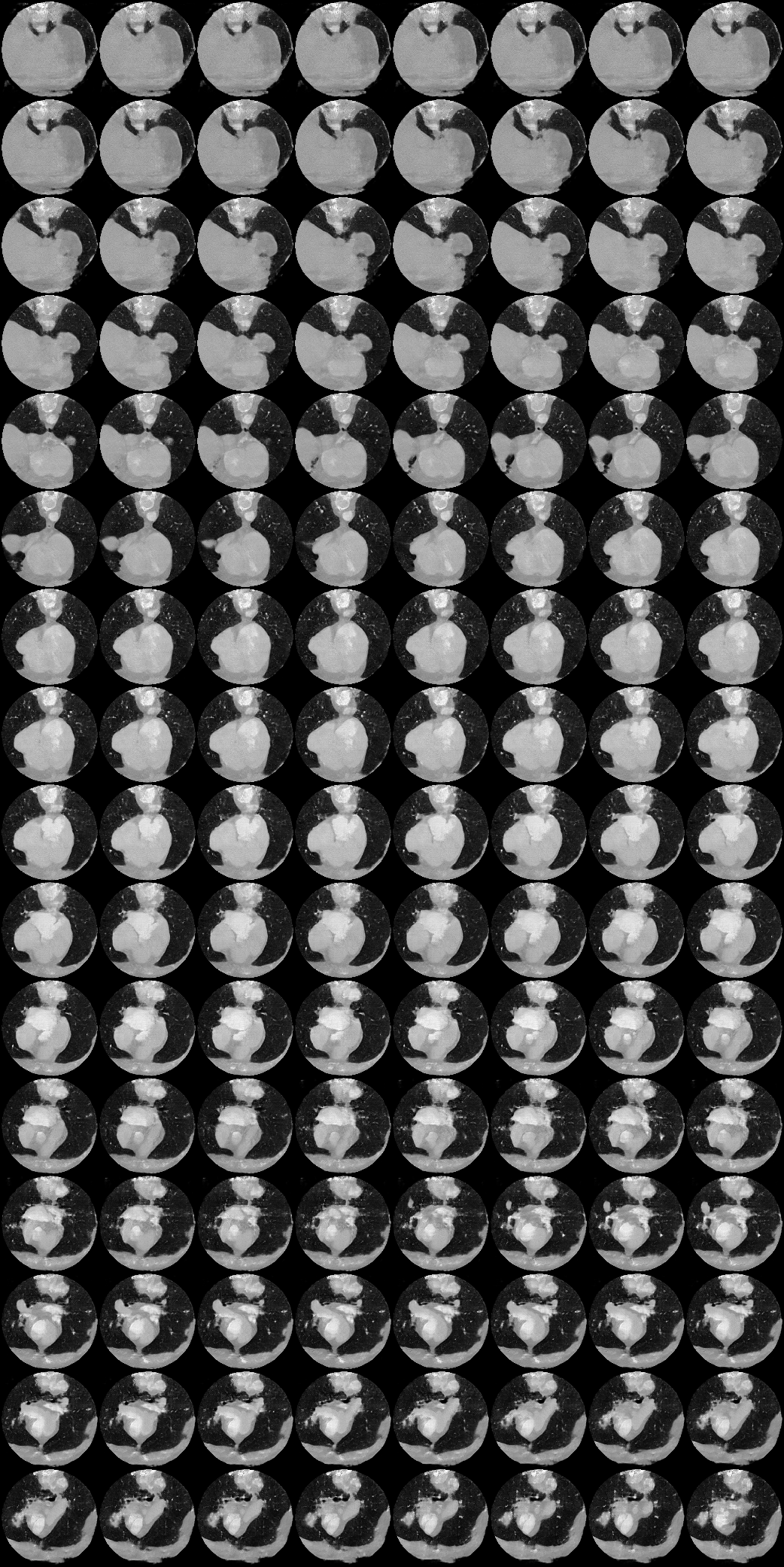}
&
\includegraphics[width=0.225\linewidth]{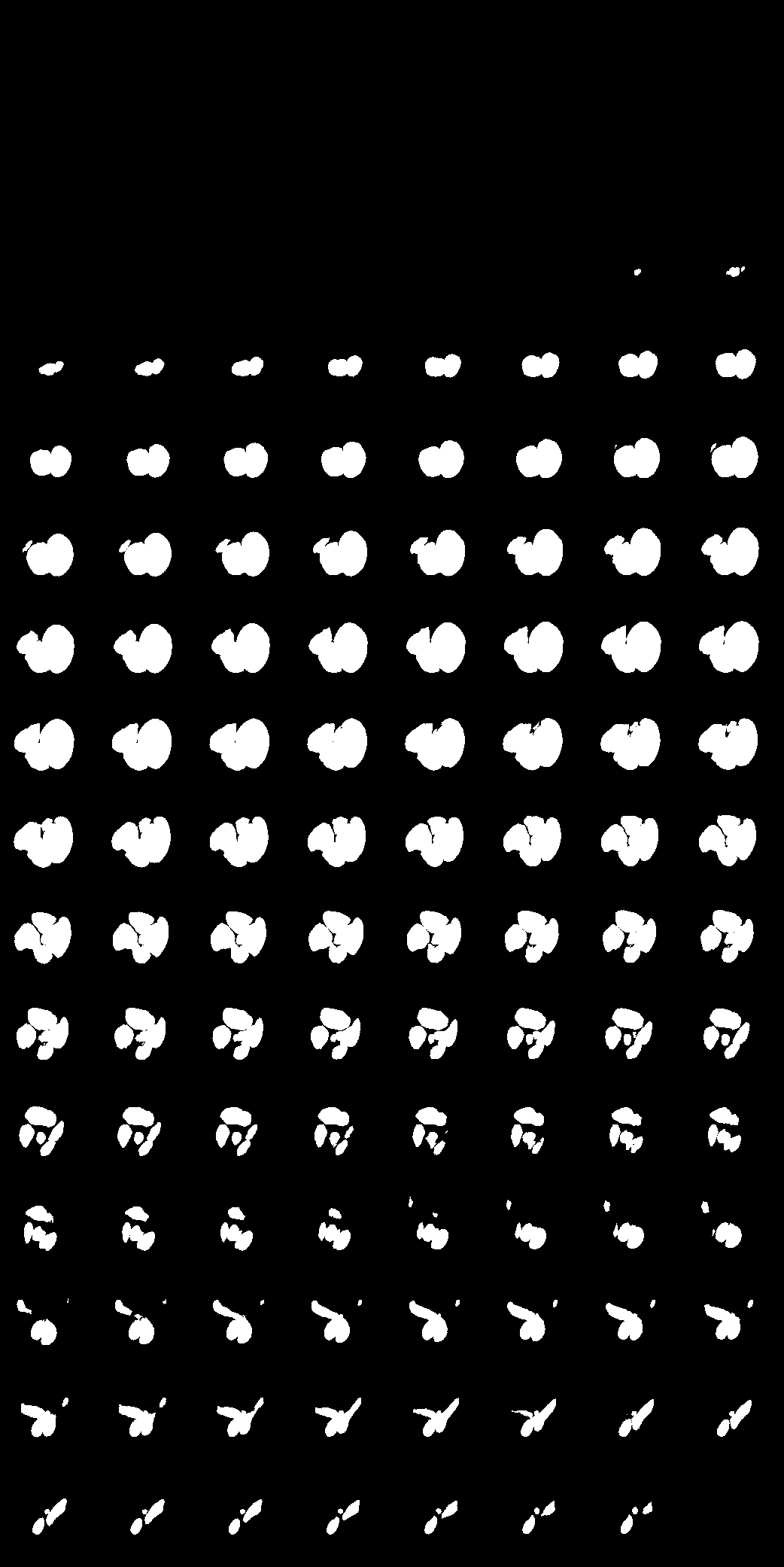}
&
\includegraphics[width=0.225\linewidth]{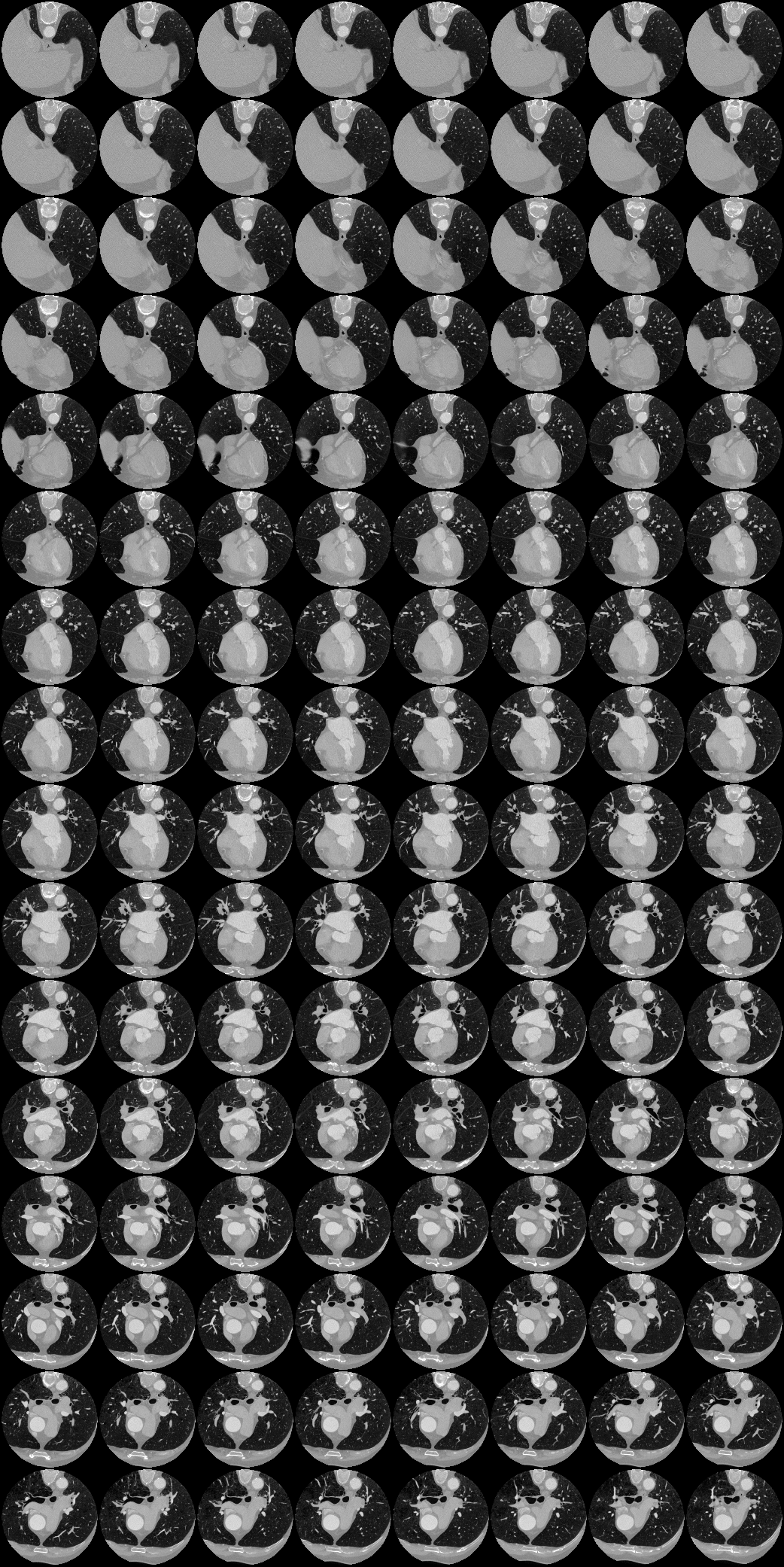}
&
\includegraphics[width=0.225\linewidth]{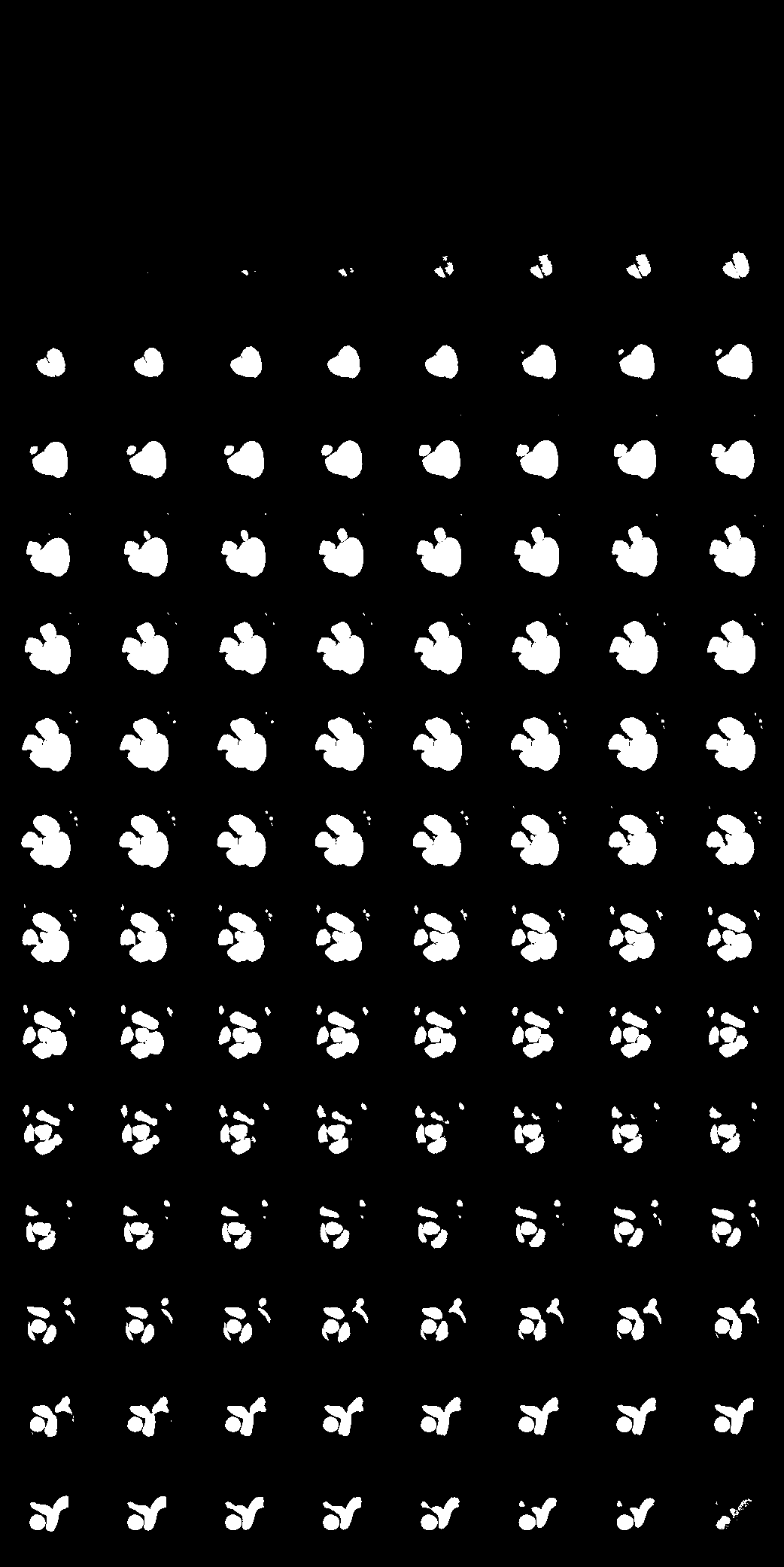}\\

\rotatebox{90}{\ {\color{black}{\scriptsize CT34LC}}} &\includegraphics[width=0.225\linewidth]{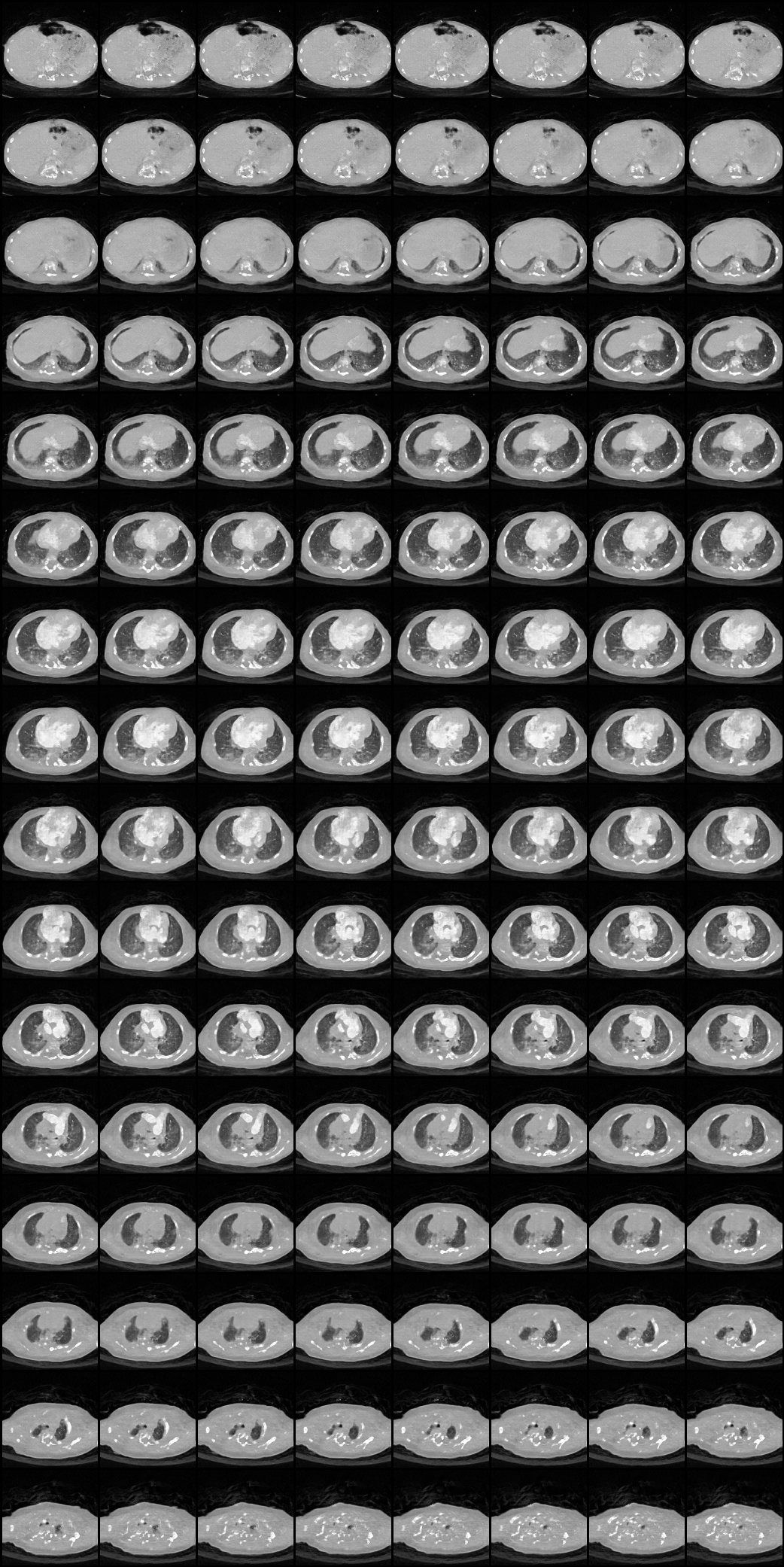}
&
\includegraphics[width=0.225\linewidth]{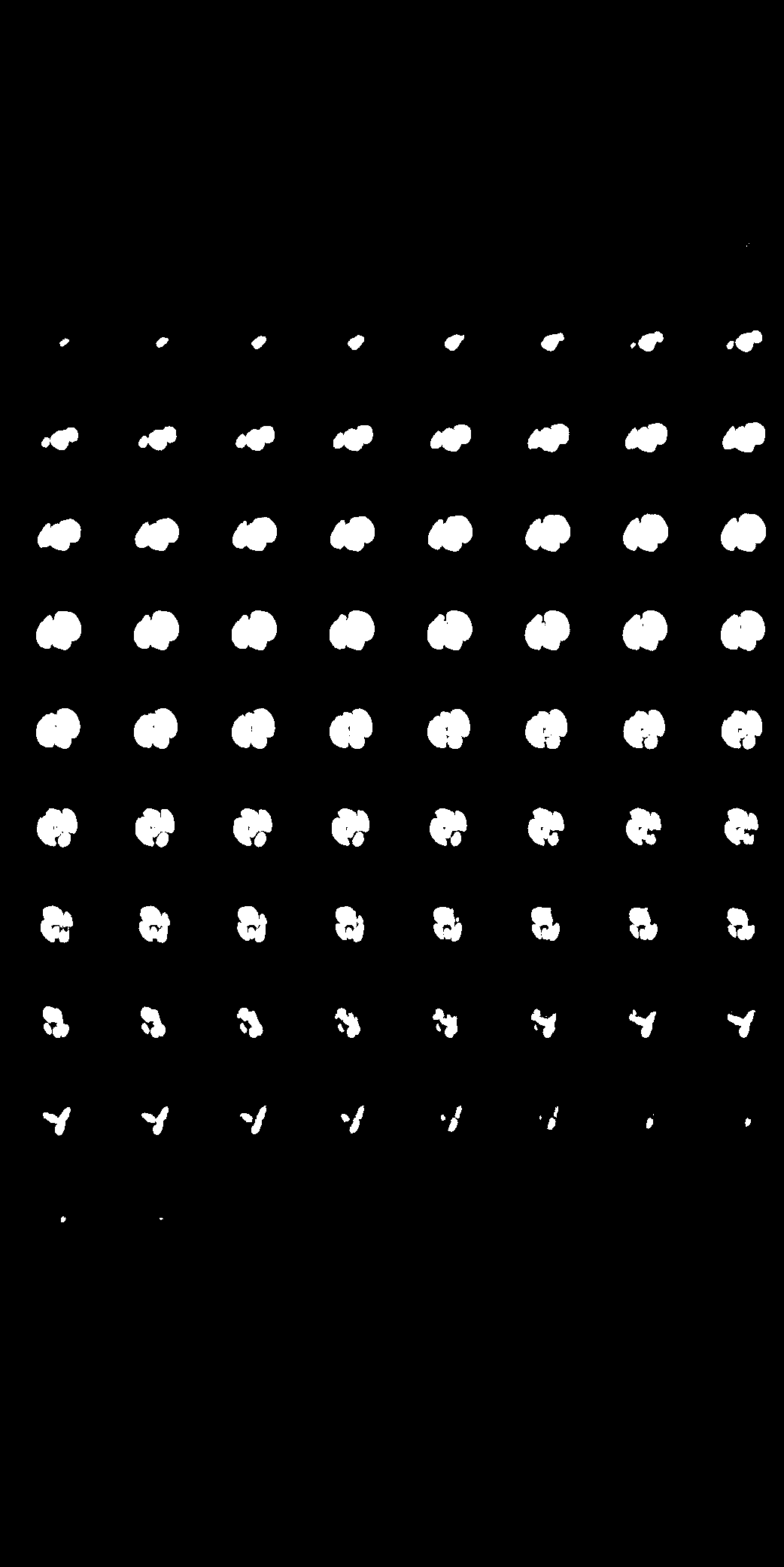}
&
\includegraphics[width=0.225\linewidth]{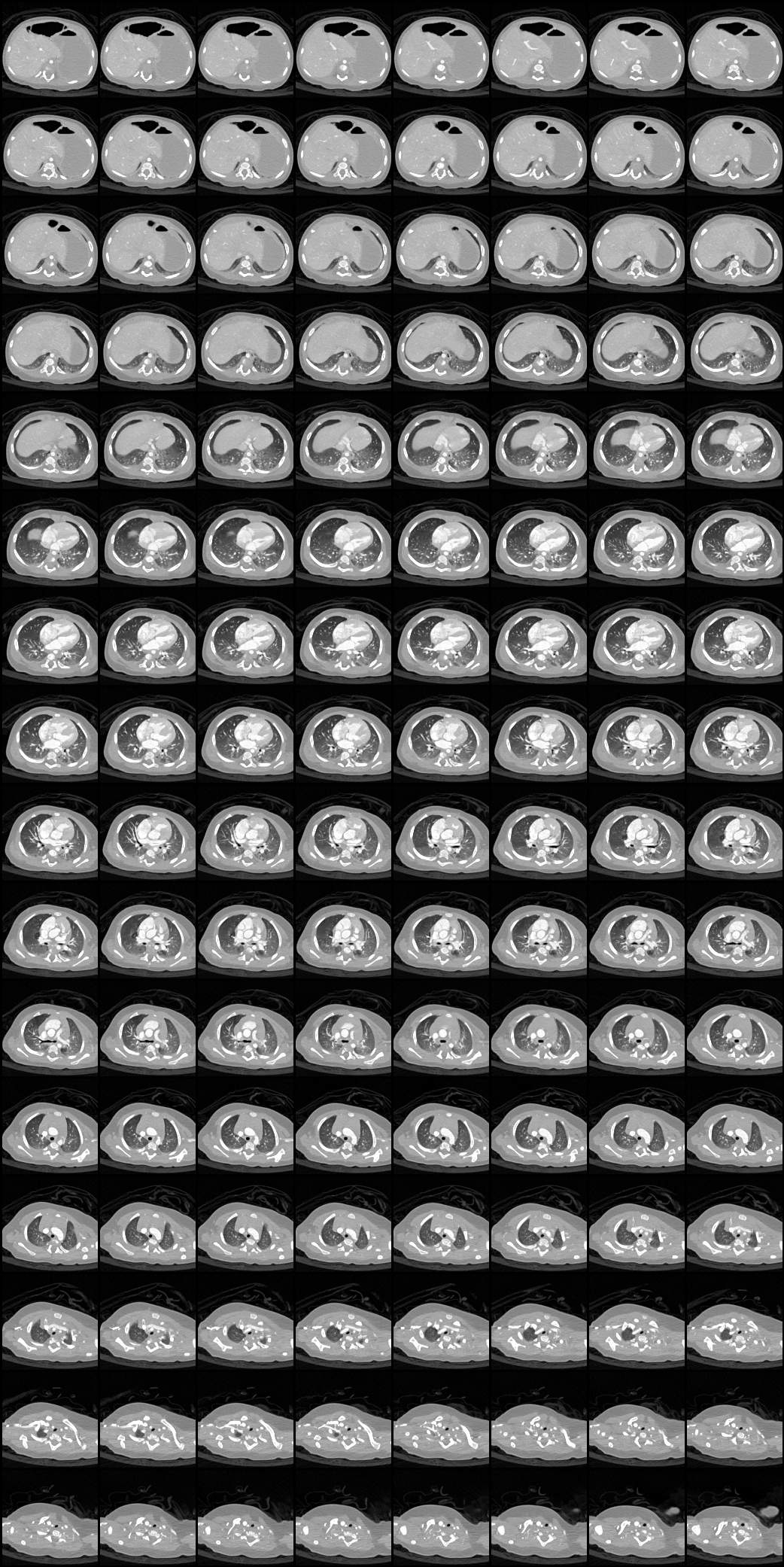}
&
\includegraphics[width=0.225\linewidth]{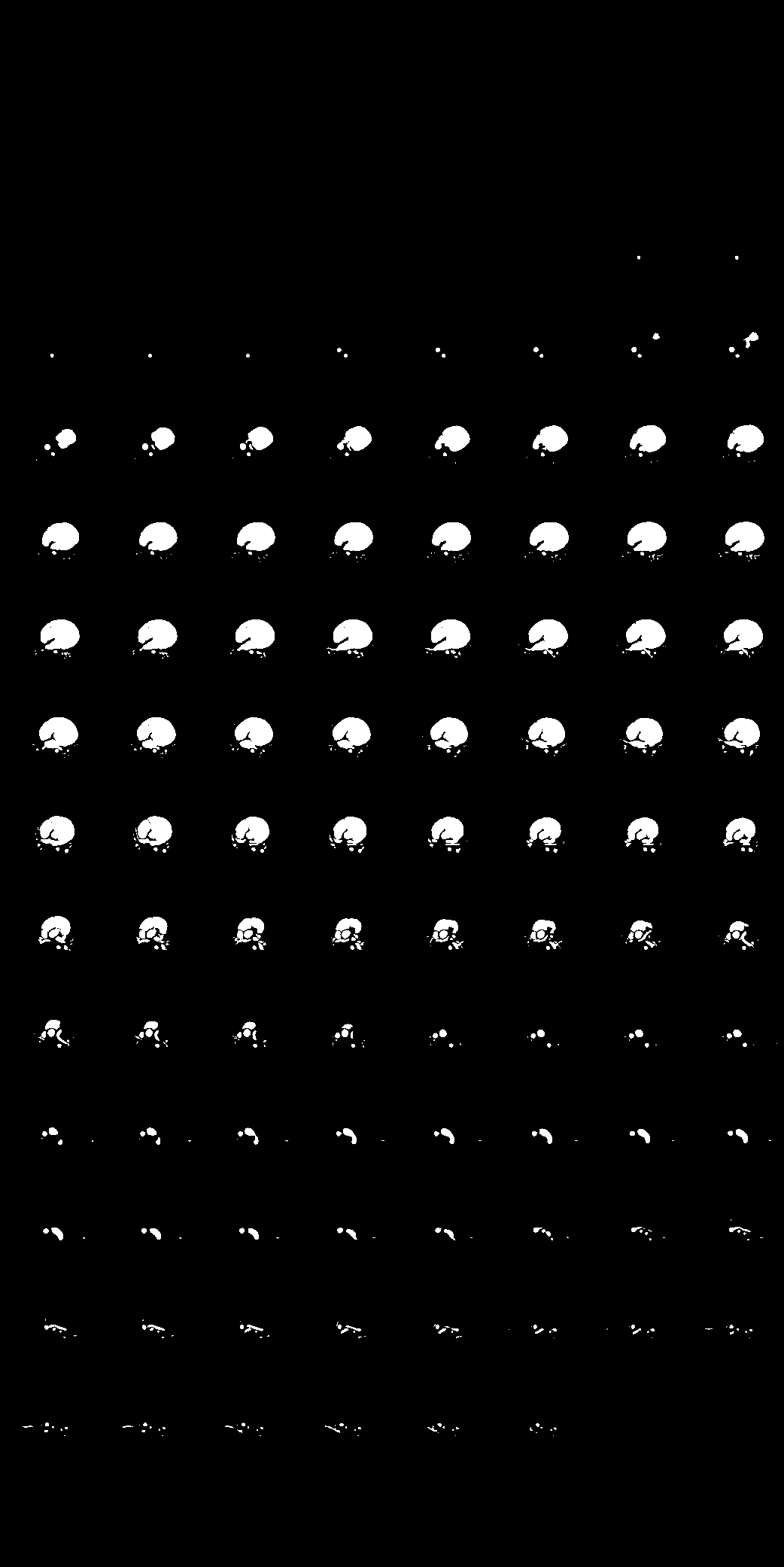}\\

\rotatebox{90}{\,{\color{black}{\scriptsize CT34MC}}} &\includegraphics[width=0.225\linewidth]{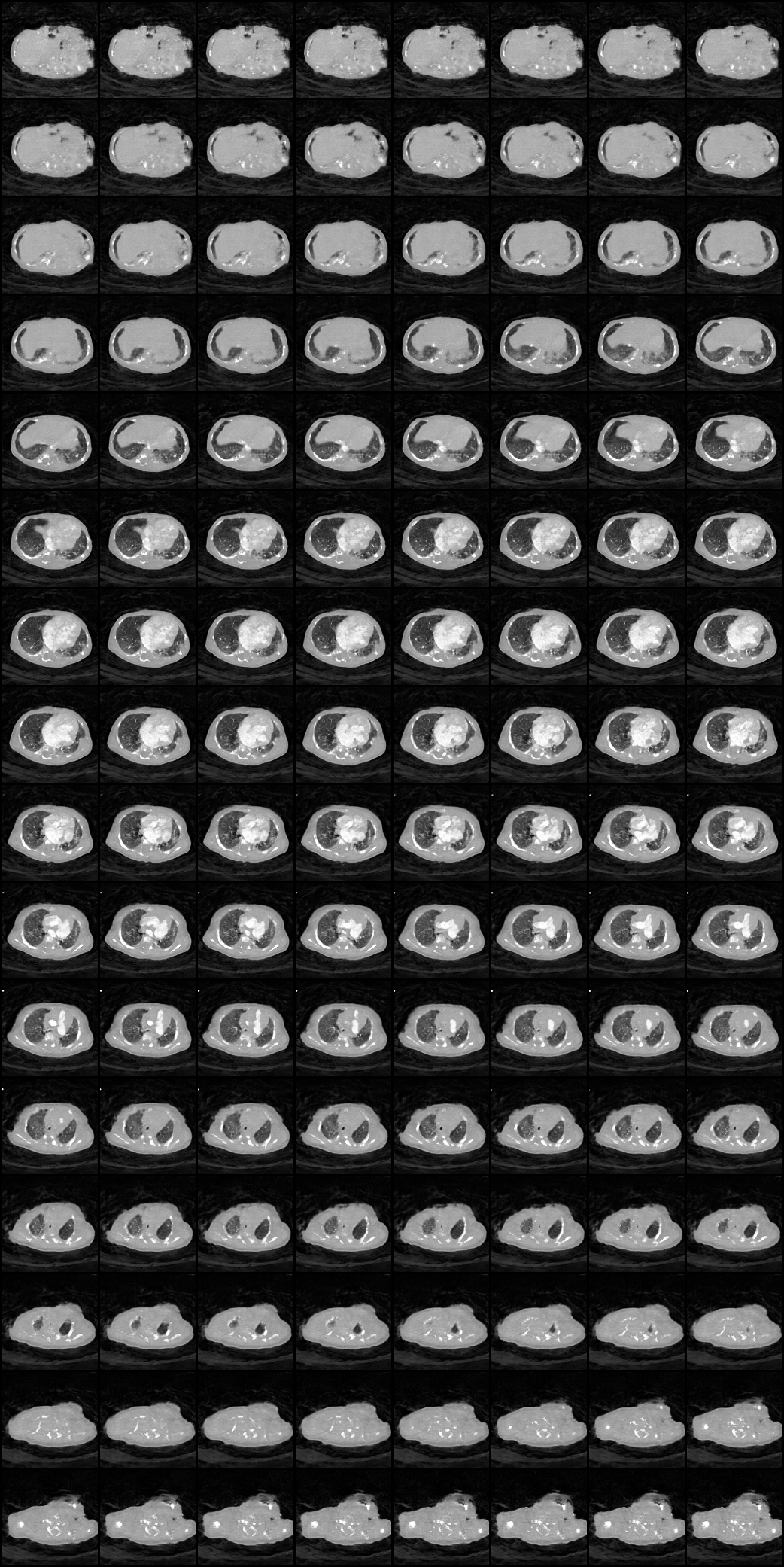}
&
\includegraphics[width=0.225\linewidth]{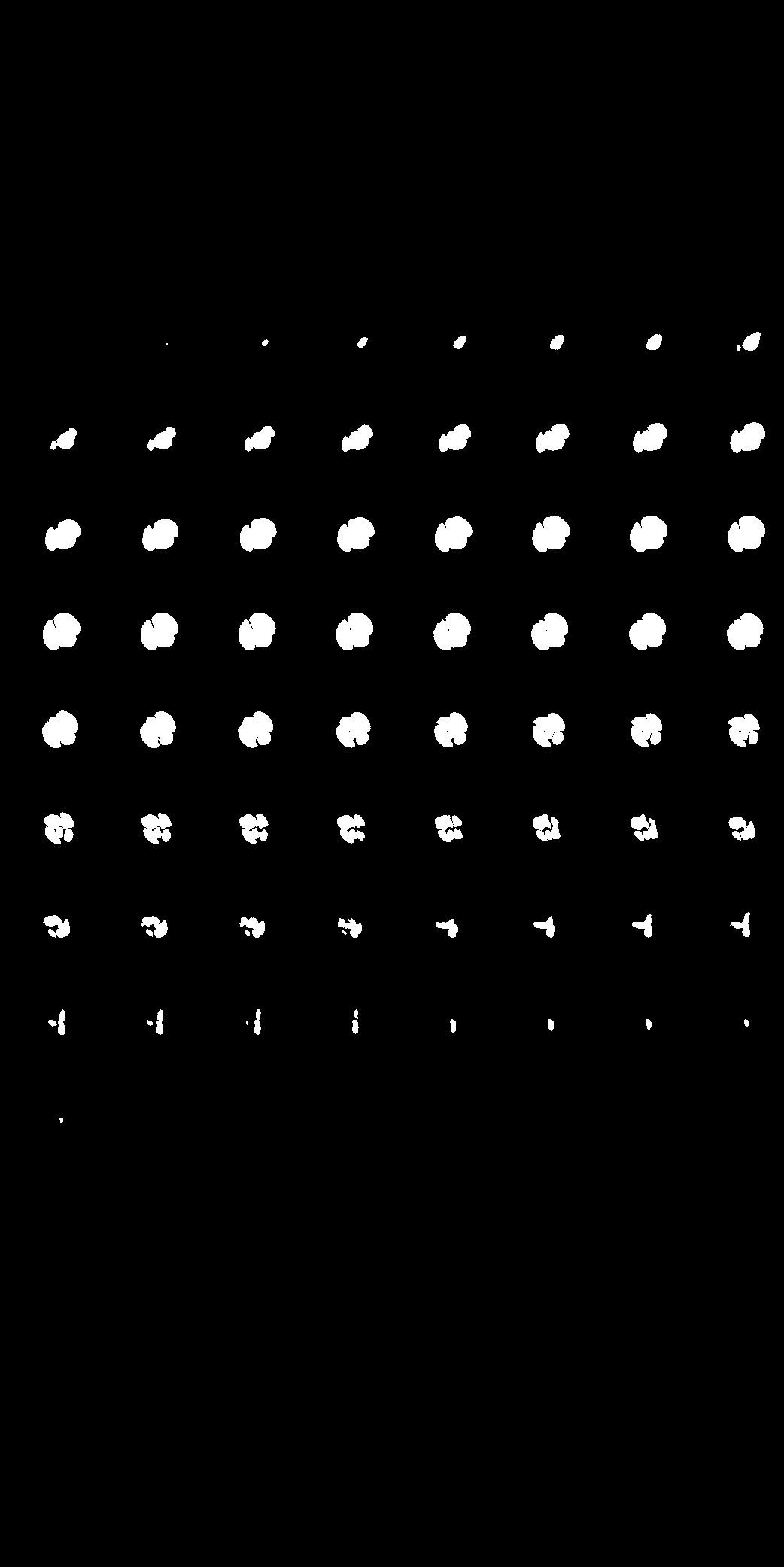}
&
\includegraphics[width=0.225\linewidth]{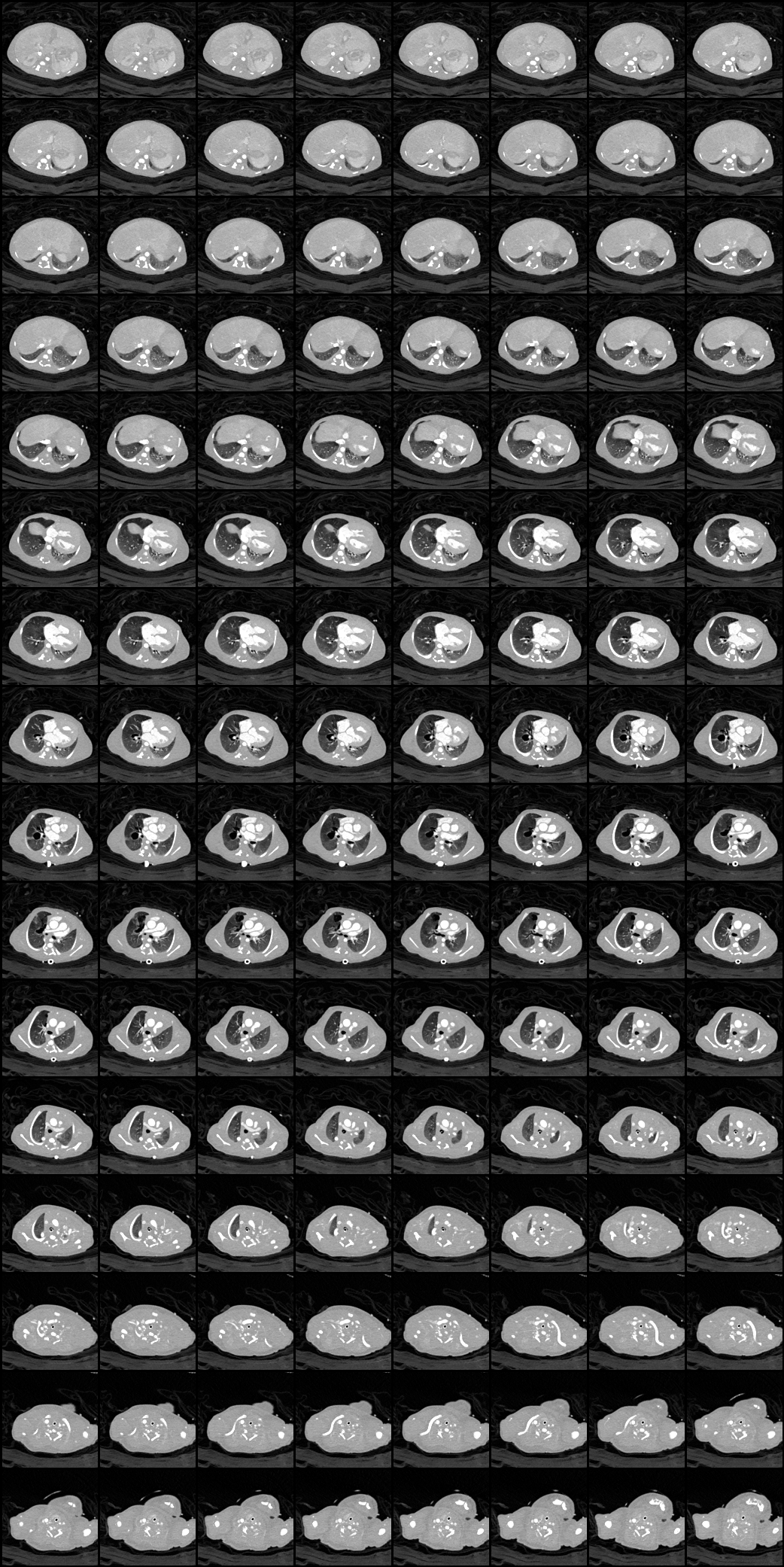}
&
\includegraphics[width=0.225\linewidth]{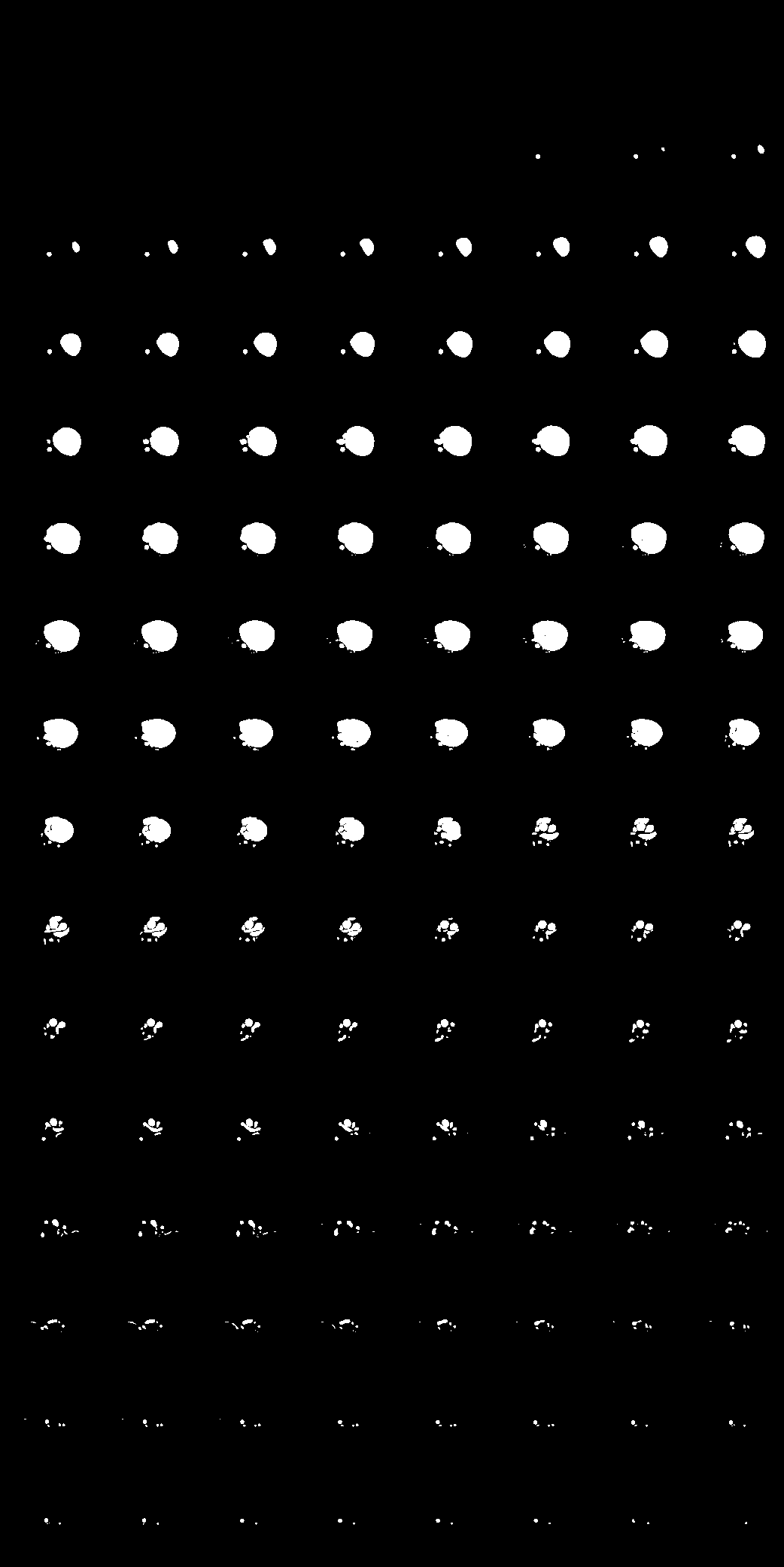}\\
\end{tabular}

\caption{Qualitative Results: First two columns show random samples (full volume) from our full model on each of the datasets. Last two columns show nearest neighbour from the training set. We see that our model can generate plausible yet novel data samples with annotations (second column).}
\label{fig:qual}
\end{figure}

\begin{figure}[t!]
\centering
\begin{tabular}{cccc}
\rotatebox{90}{\ \ \,{\color{black}{\scriptsize Ours-Fix-Mat}}} &\includegraphics[width=0.25\linewidth]{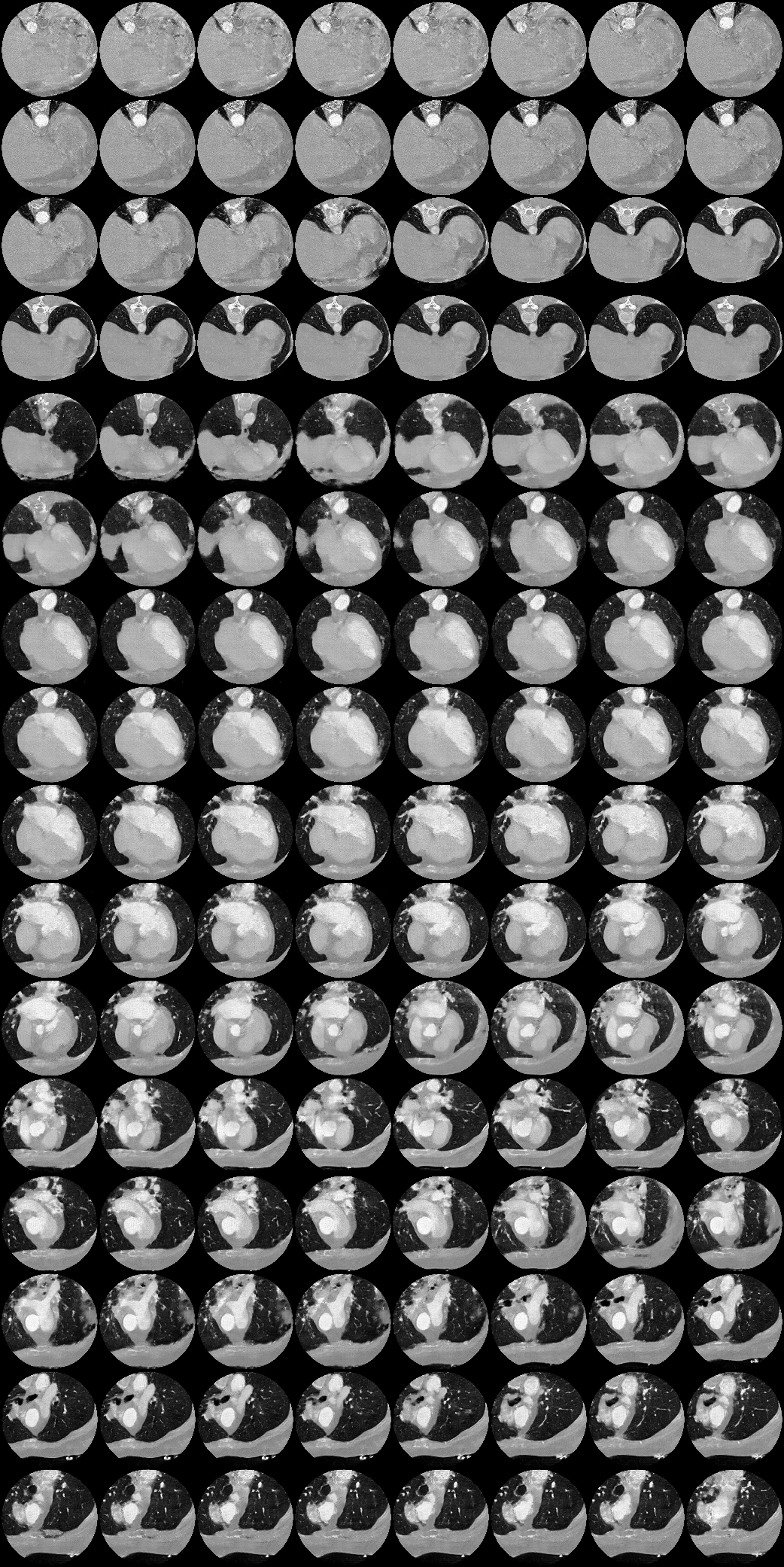}
&
\includegraphics[width=0.25\linewidth]{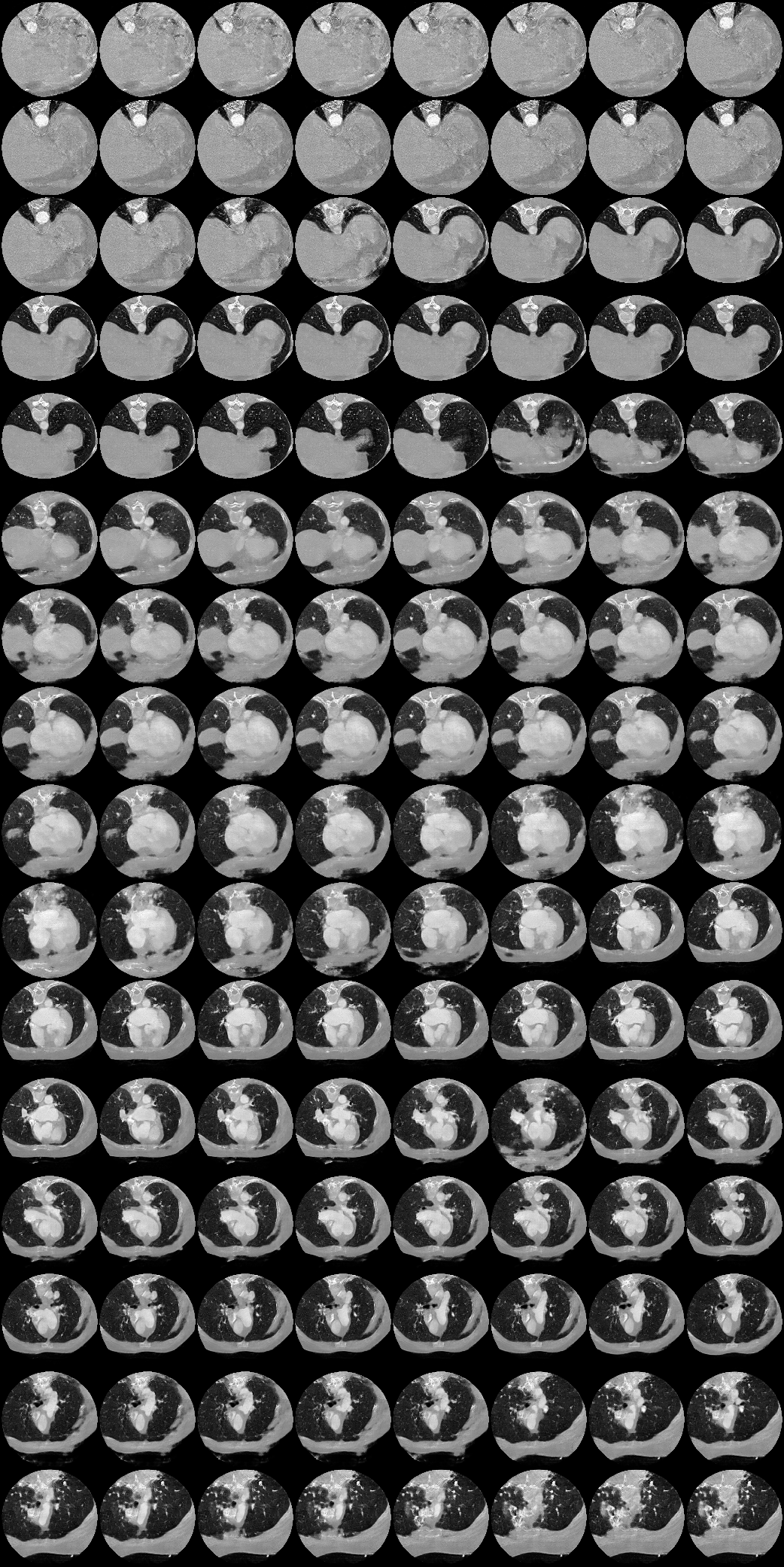}
&
\includegraphics[width=0.25\linewidth]{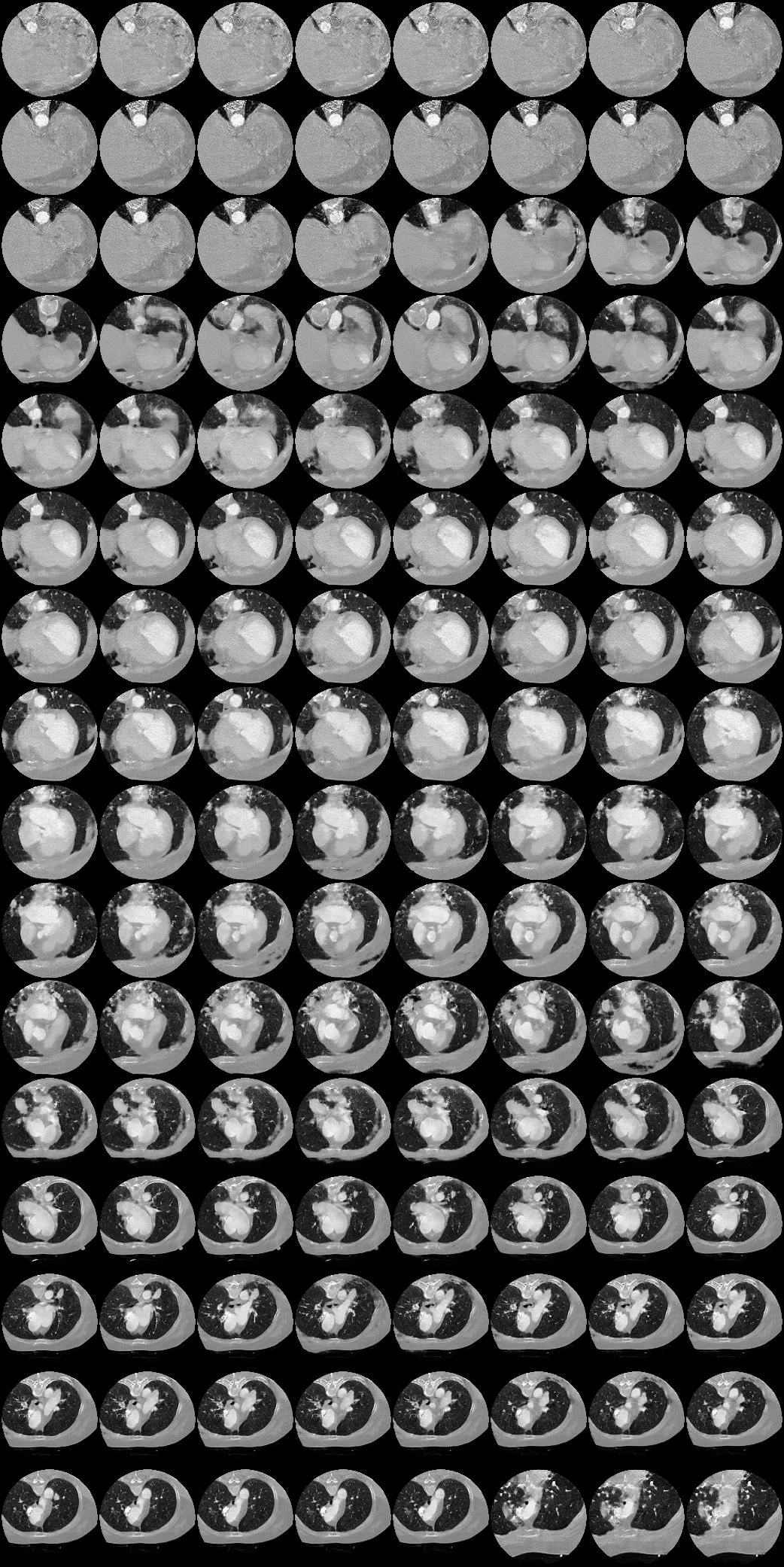}\\

\rotatebox{90}{\ {\color{black}{\scriptsize Ours-Pre}}} &\includegraphics[width=0.25\linewidth]{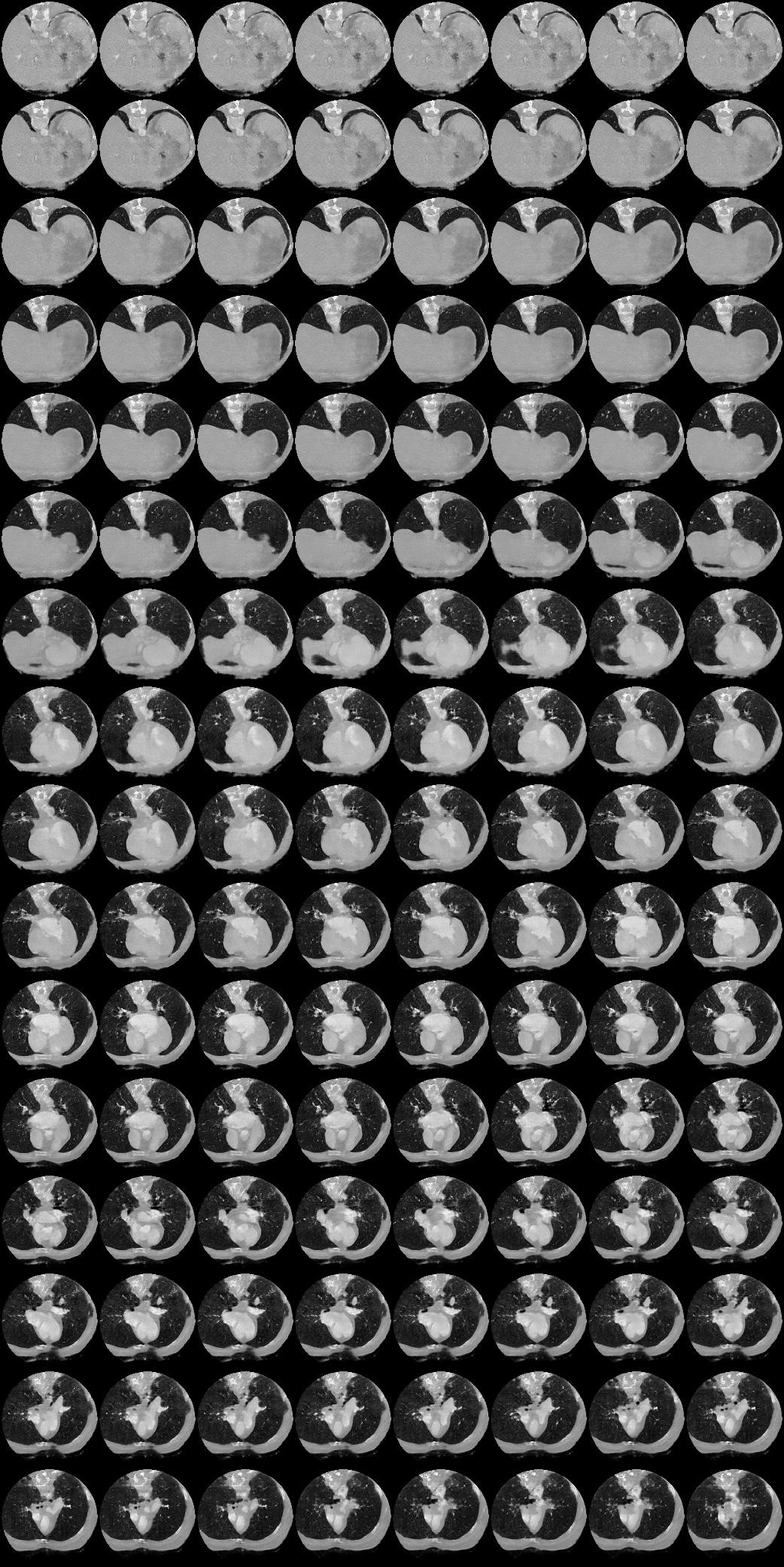}
&
\includegraphics[width=0.25\linewidth]{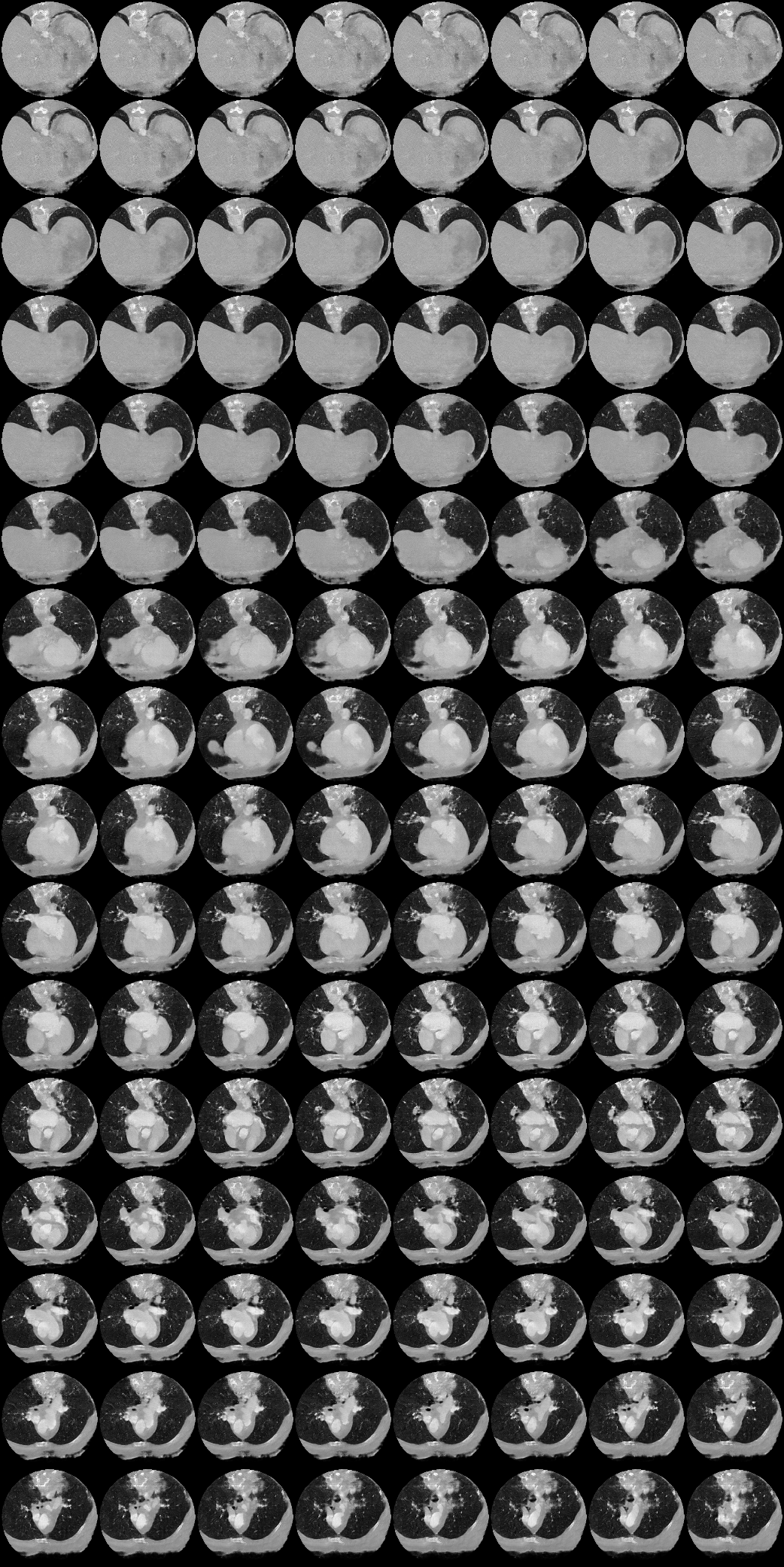}
&
\includegraphics[width=0.25\linewidth]{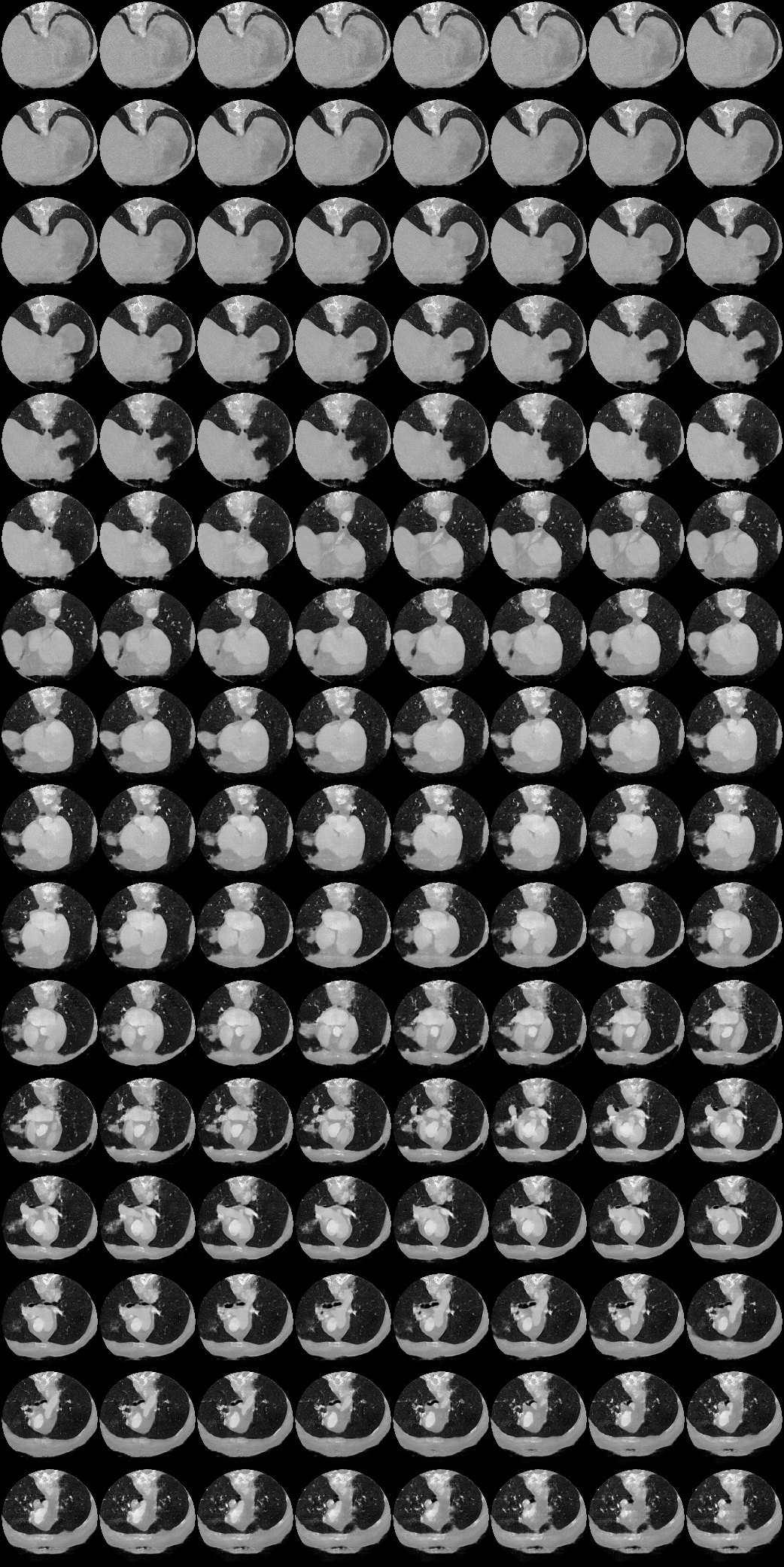}\\

\rotatebox{90}{\,{\color{black}{\scriptsize Ours-Full}}} &\includegraphics[width=0.25\linewidth]{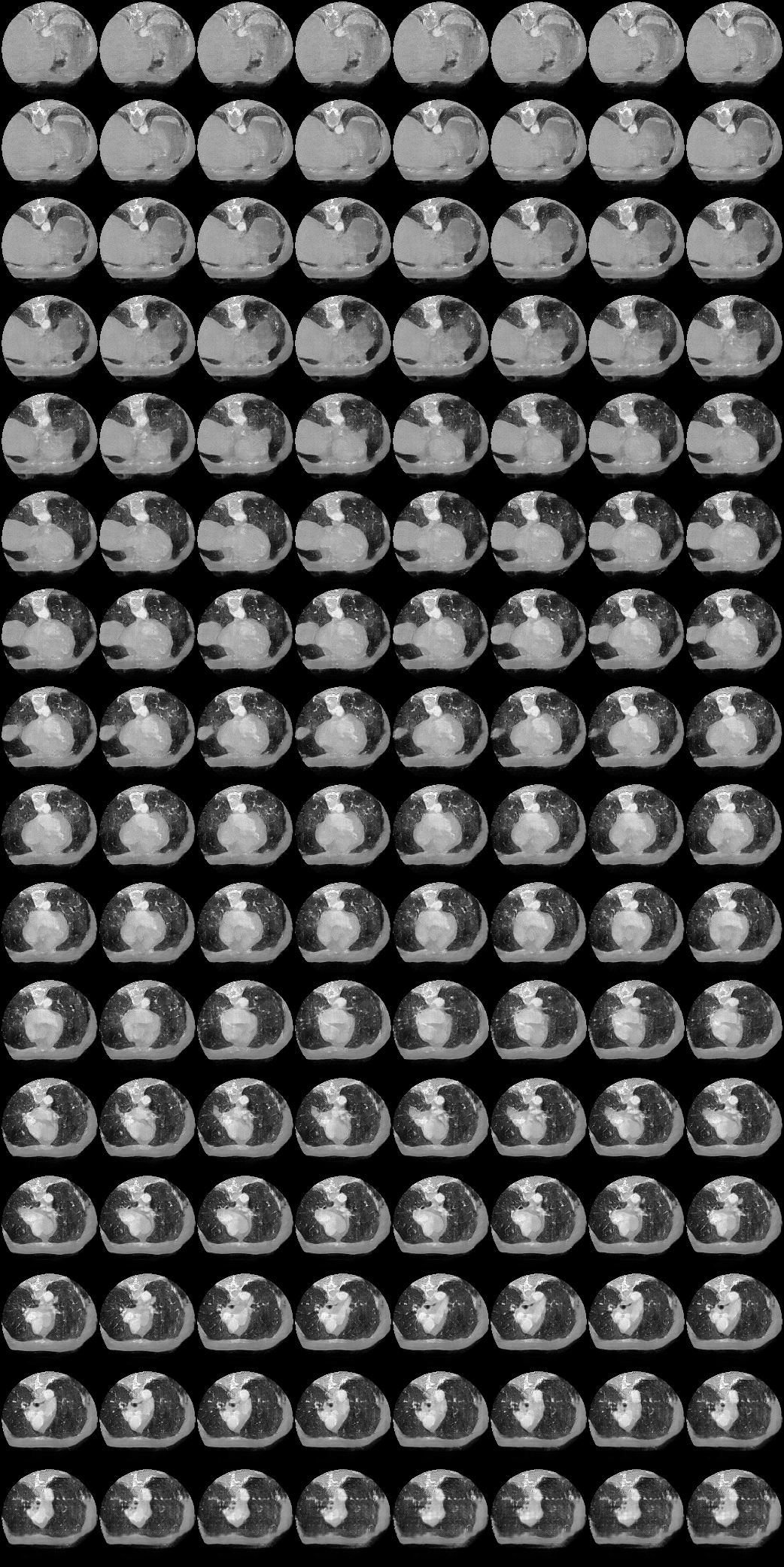}
&
\includegraphics[width=0.25\linewidth]{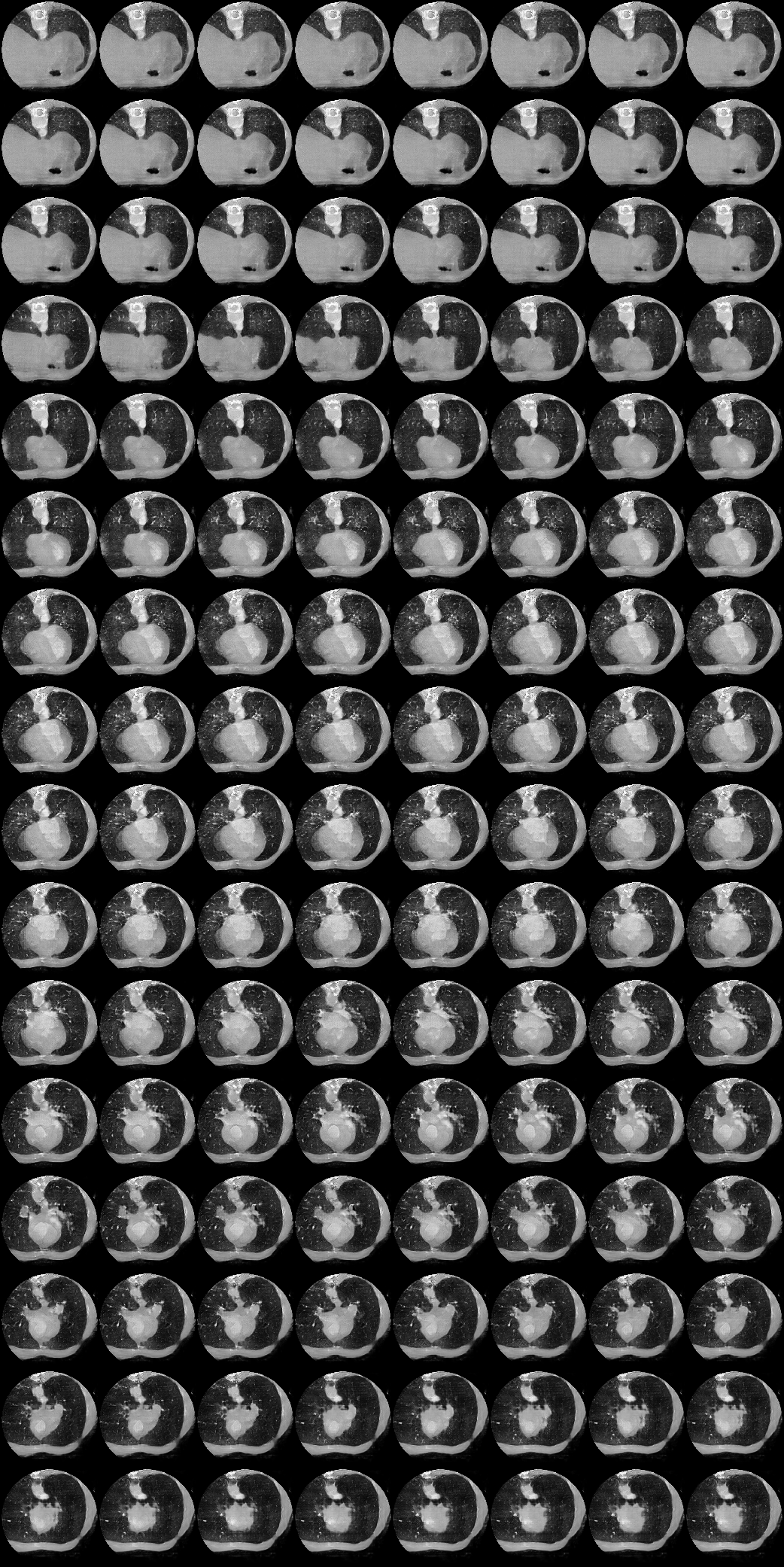}
&
\includegraphics[width=0.25\linewidth]{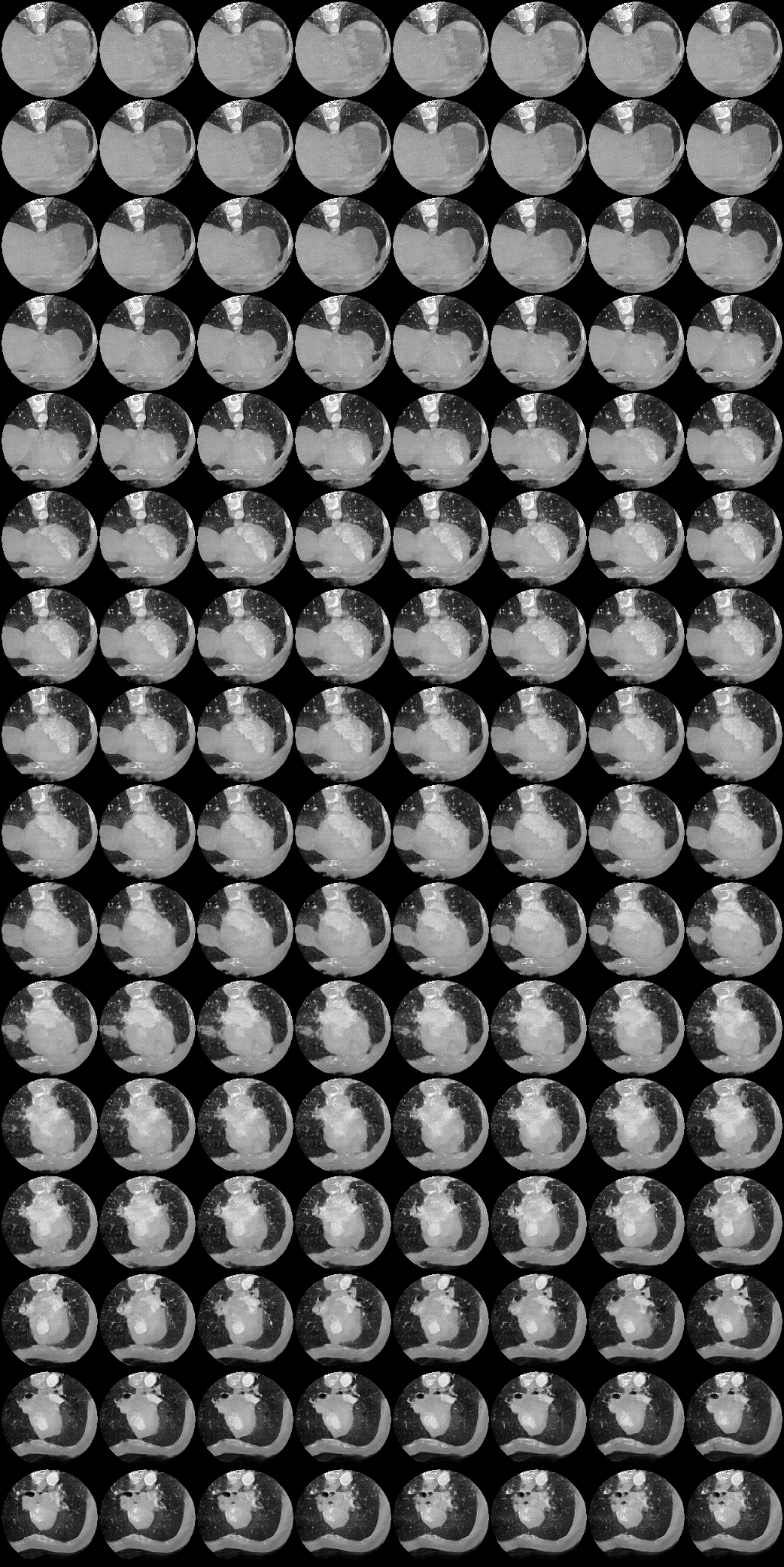}\\
\end{tabular}

\caption{Qualitative Results: Three random samples(Full Volume) from our different methods. Ours-Pre has better image quality and slice consistency comparing to Ours-Fix-Mat. Ours-Full has more diverse object shape and background comparing to the other two methods.}
\label{fig:qual}
\end{figure}

\begin{figure}[t!]
\centering
\begin{tabular}{cccc}
\rotatebox{90}{\ \ \,{\color{black}{\scriptsize CT20}}} &\includegraphics[width=0.25\linewidth]{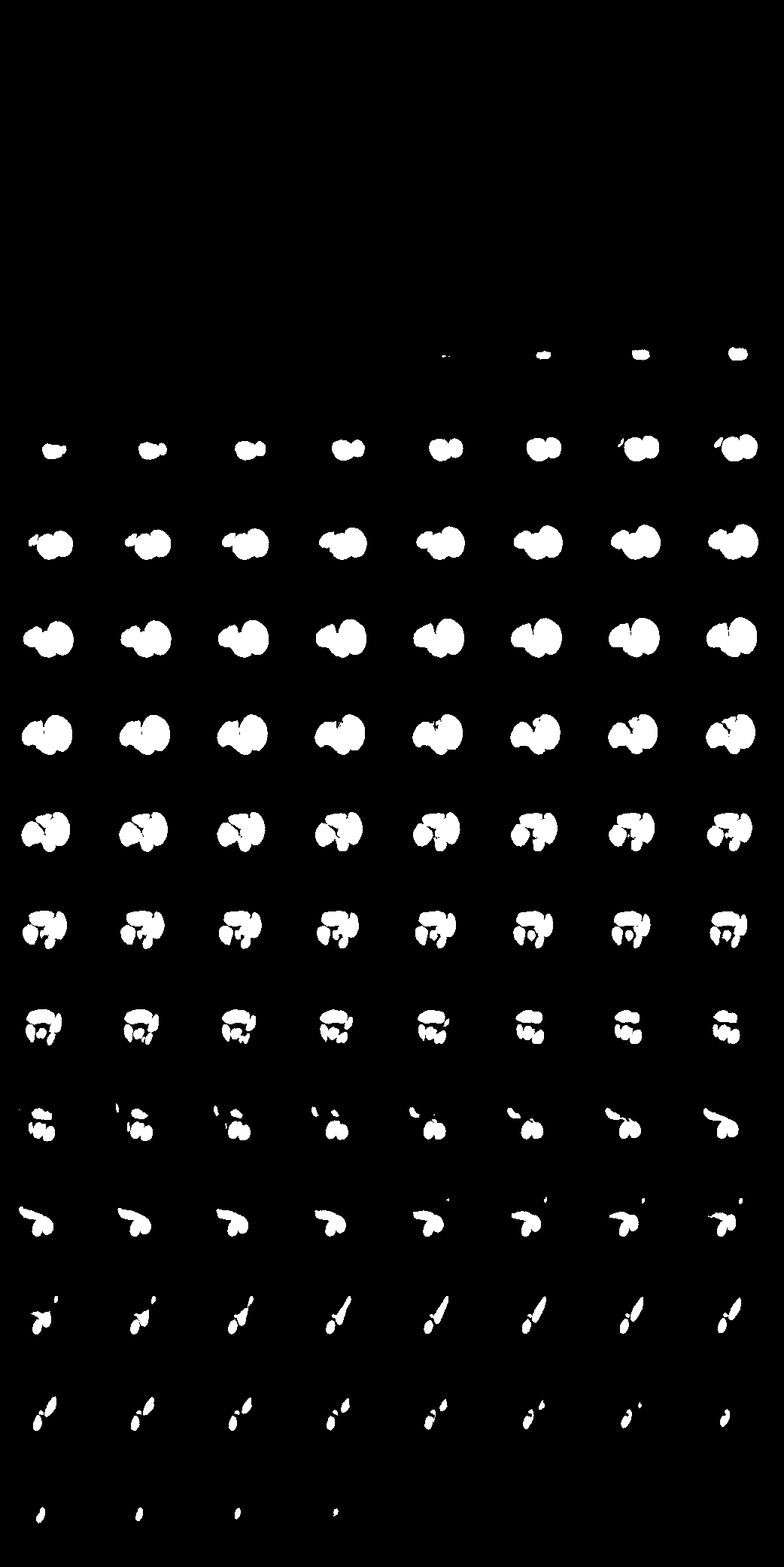}
&
\includegraphics[width=0.25\linewidth]{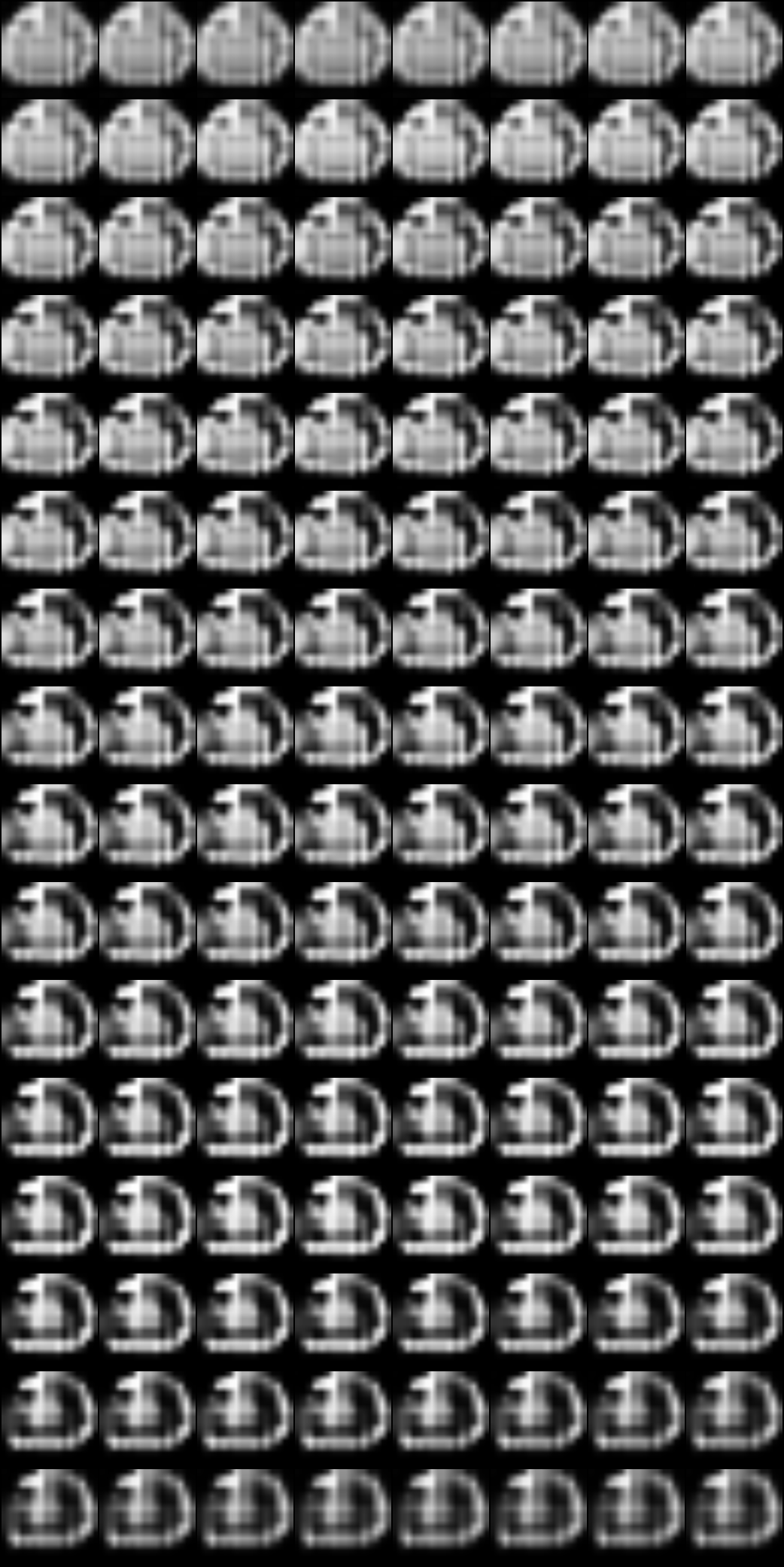}
&
\includegraphics[width=0.25\linewidth]{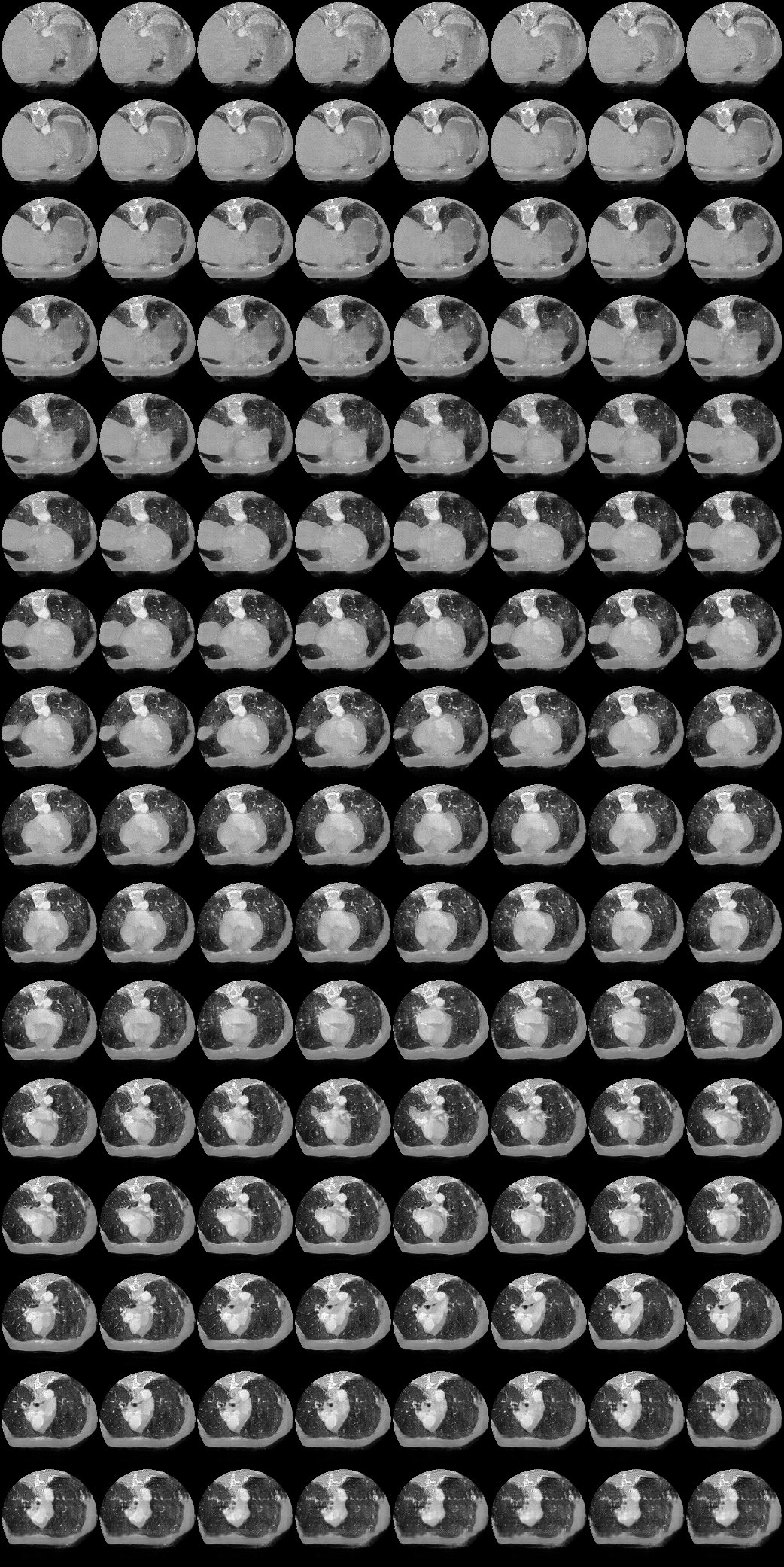}\\

\rotatebox{90}{\ {\color{black}{\scriptsize CT34LC}}} &\includegraphics[width=0.25\linewidth]{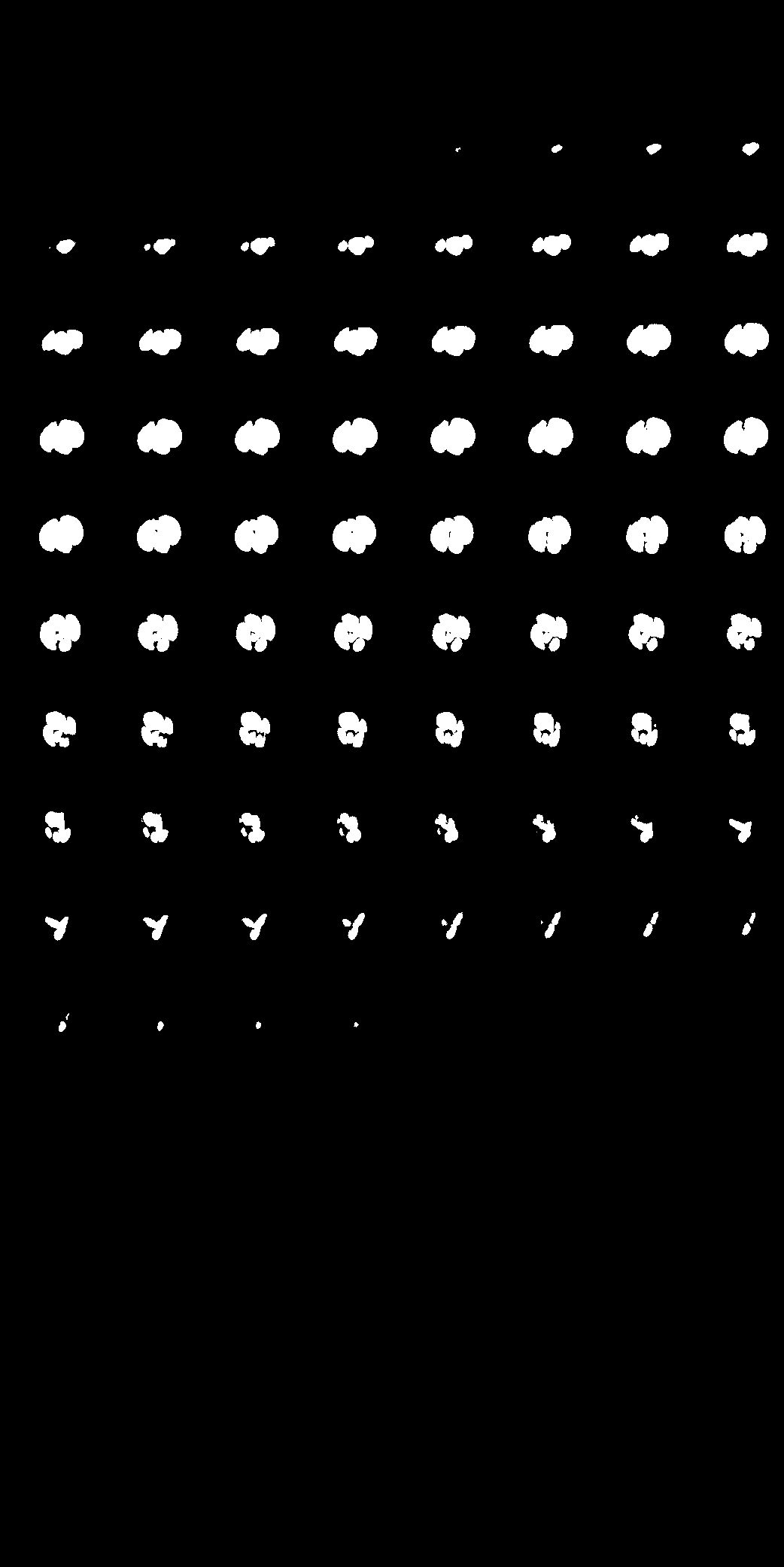}
&
\includegraphics[width=0.25\linewidth]{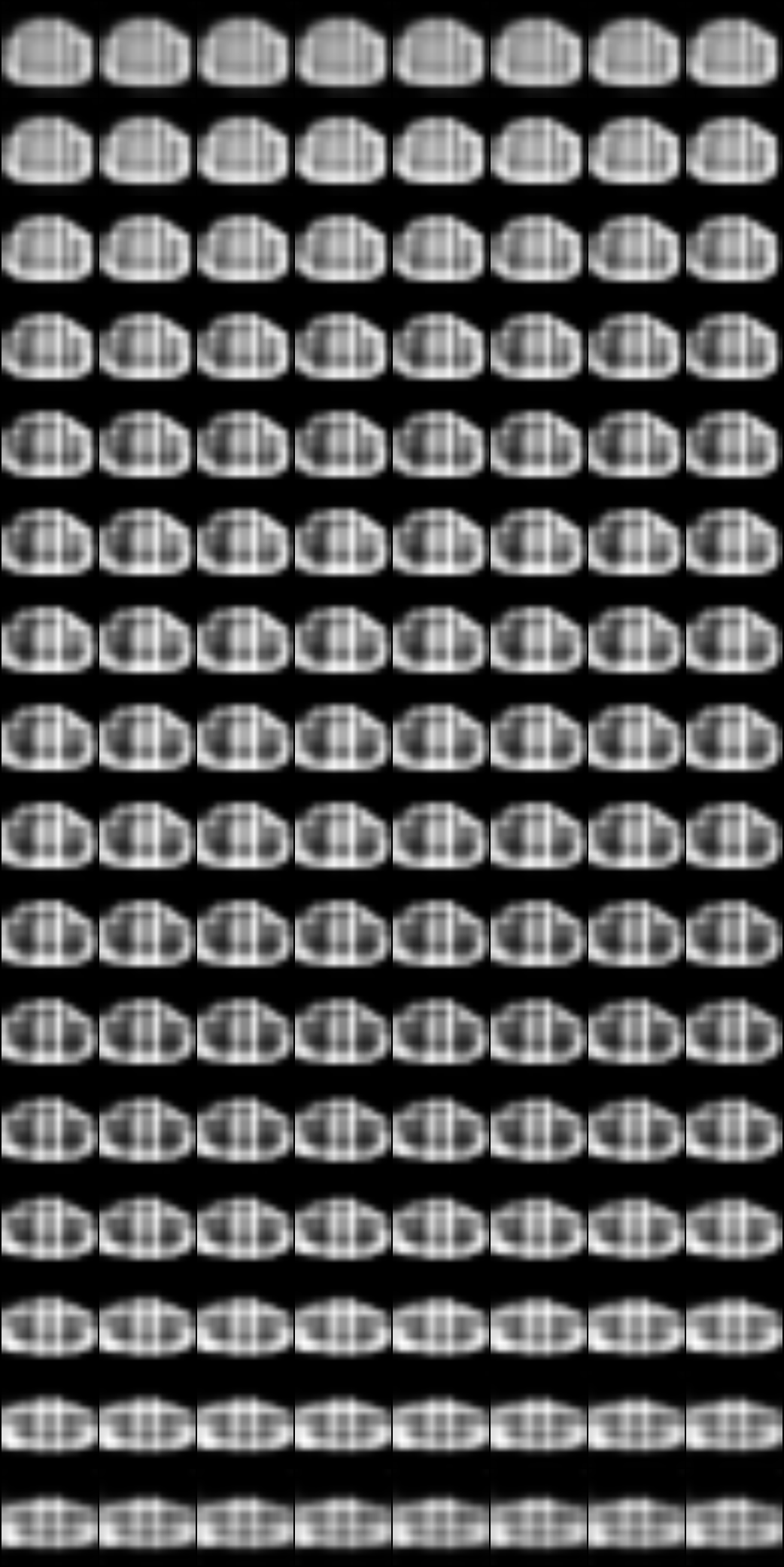}
&
\includegraphics[width=0.25\linewidth]{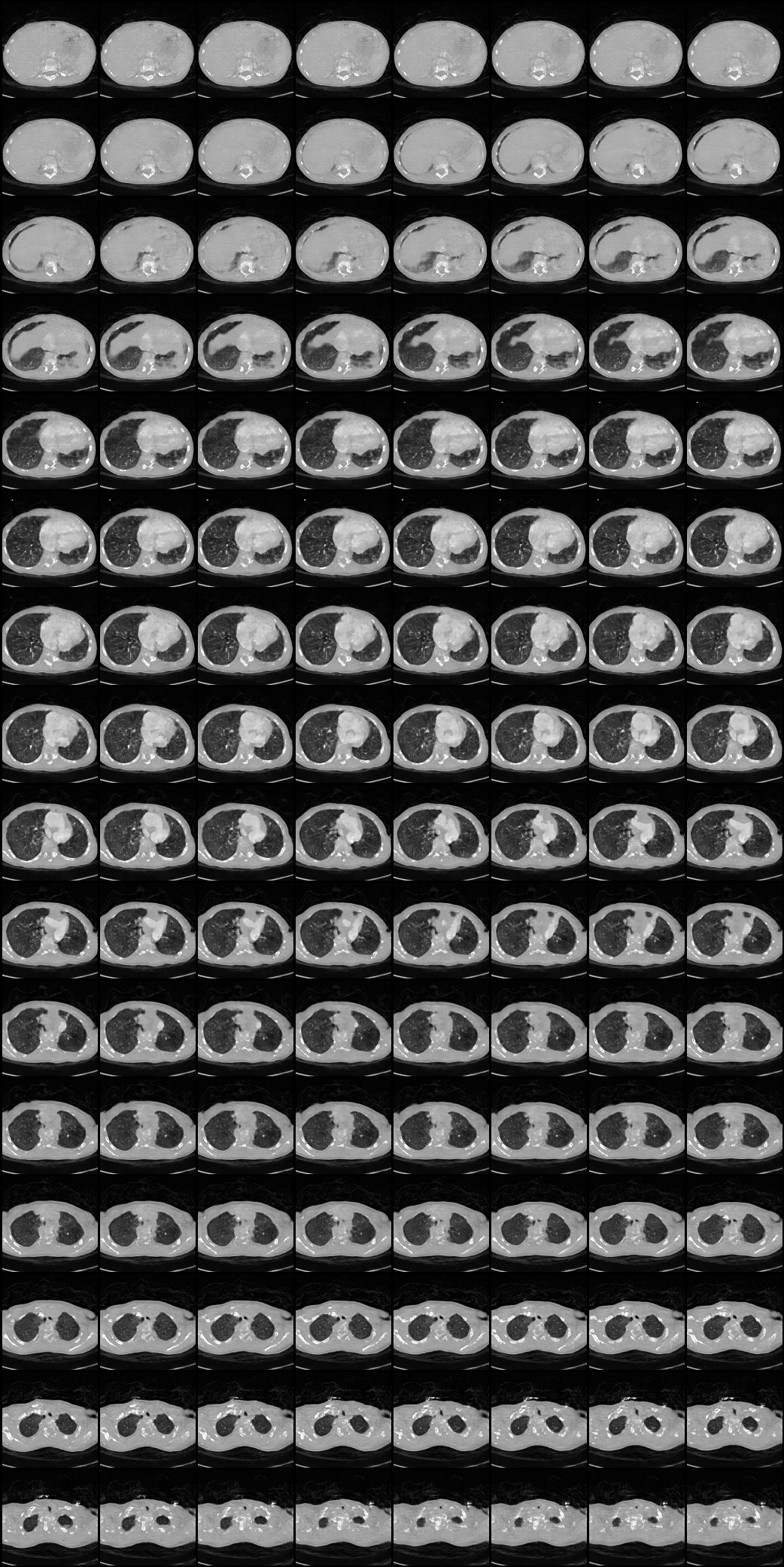}\\

\rotatebox{90}{\,{\color{black}{\scriptsize CT34MC}}} &\includegraphics[width=0.25\linewidth]{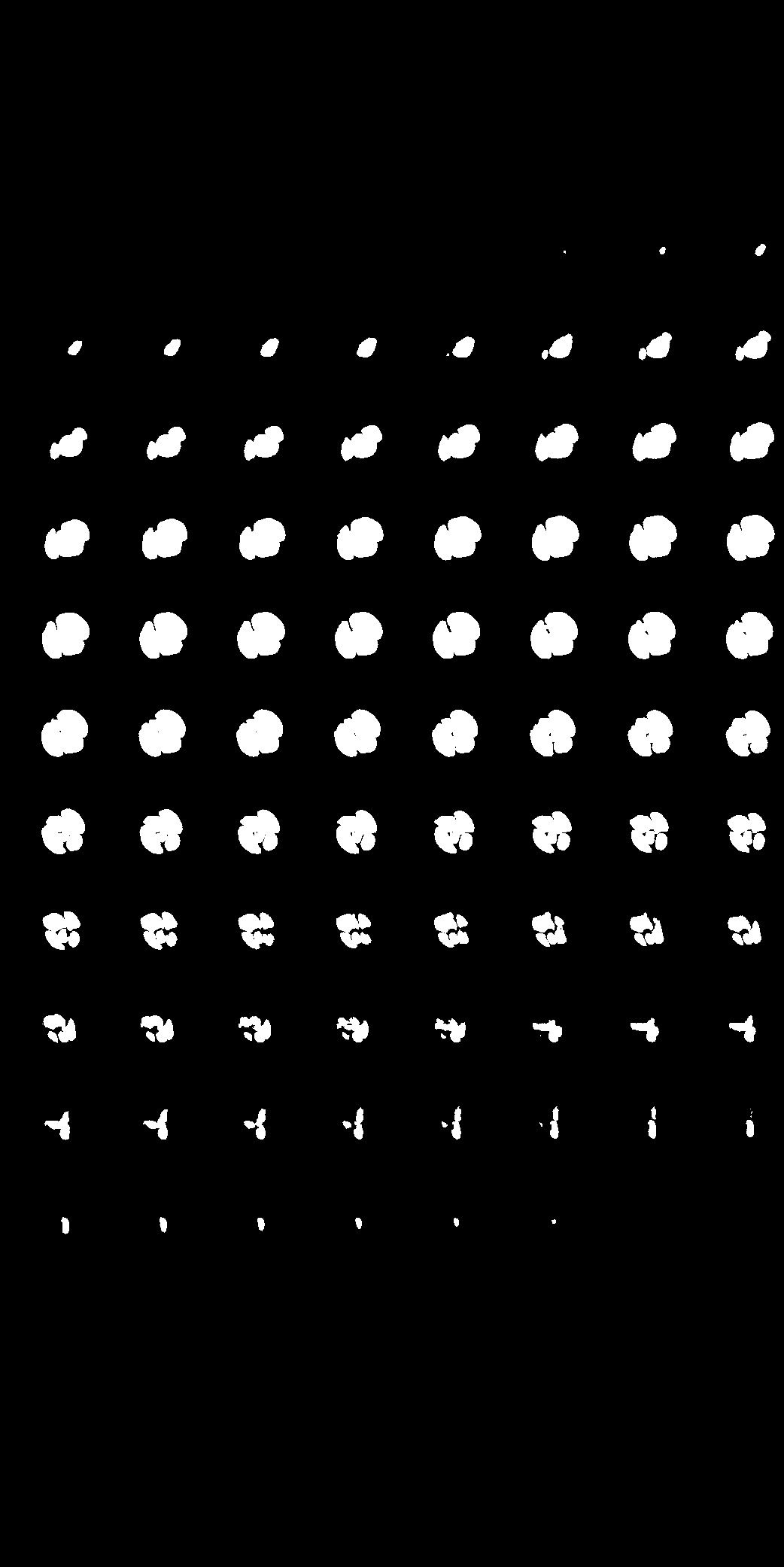}
&
\includegraphics[width=0.25\linewidth]{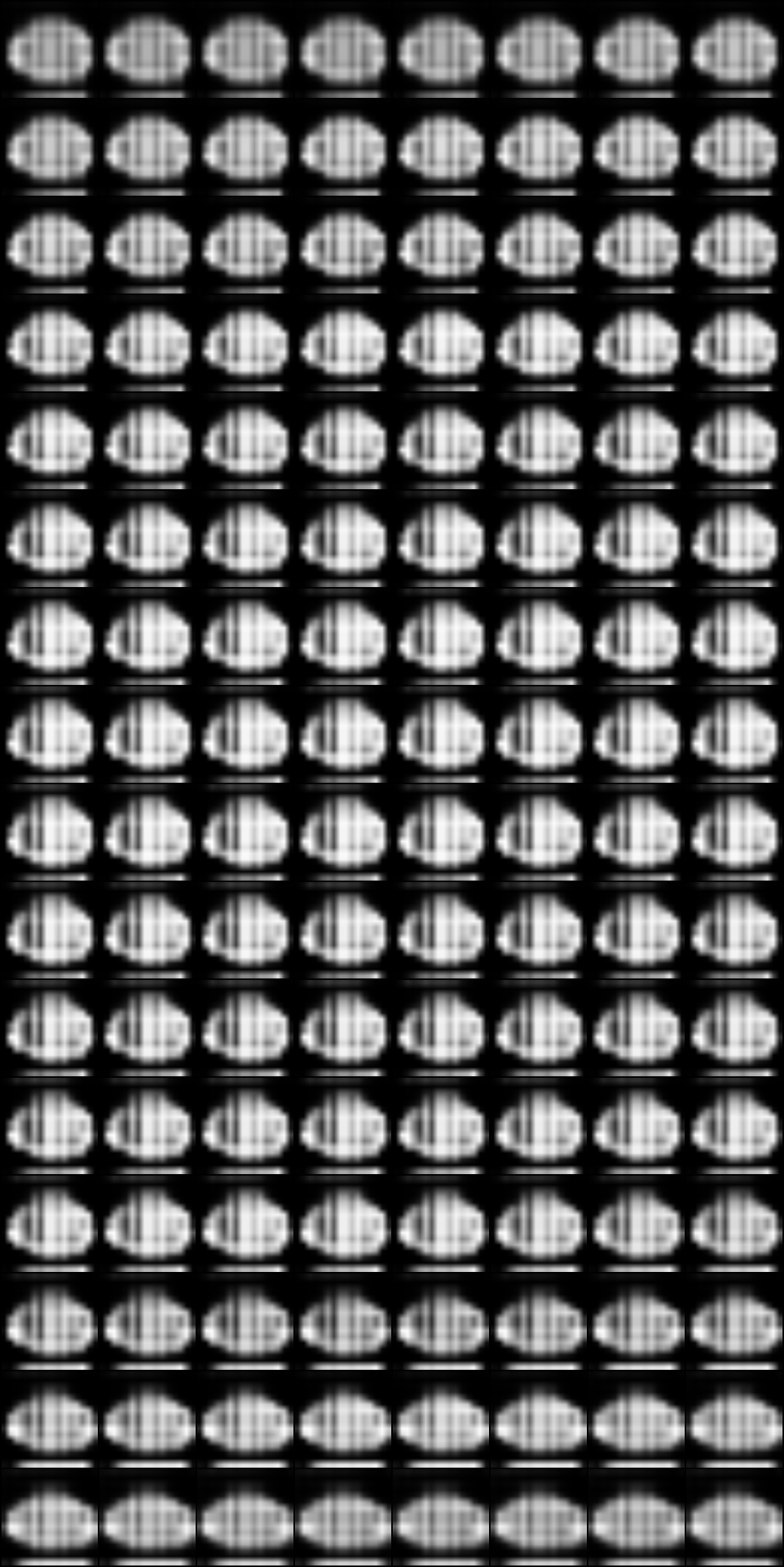}
&
\includegraphics[width=0.25\linewidth]{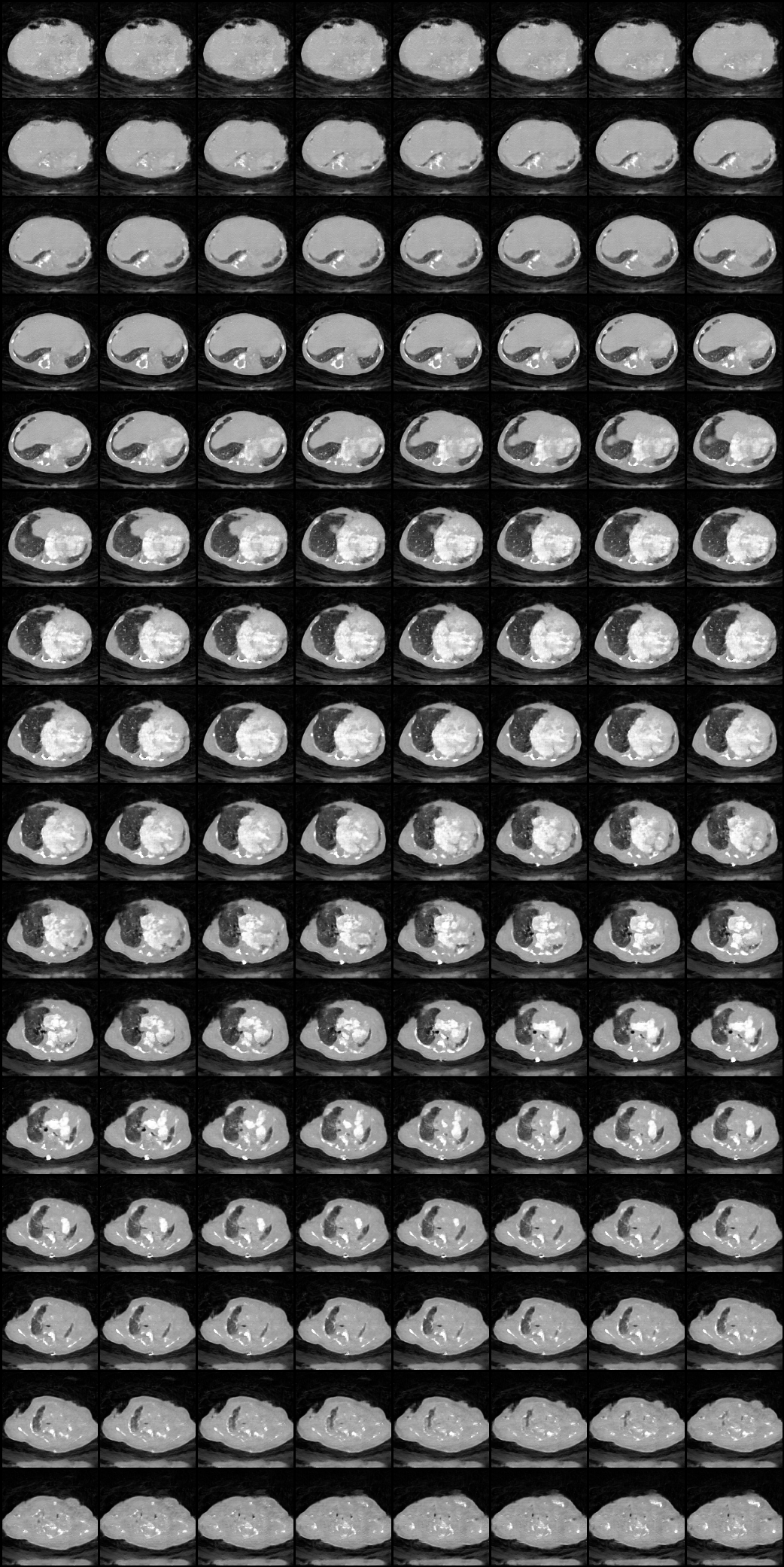}\\
\end{tabular}

\caption{Qualitative Results: Three samples generated by our method in different datasets. The first column is the generated shape, the second column is the generated material, the last column is the generated image. Note how the shape and material are consistent and correlated.}
\label{fig:qual}
\end{figure}

%
%
%
\clearpage 
\bibliographystyle{splncs04}
\bibliography{mybibliography}